\pdfoutput=1

\documentclass[11pt]{article}

\usepackage[final]{acl}

\usepackage{listings}
\usepackage{amsmath}
\usepackage{times}
\usepackage{booktabs}
\usepackage{latexsym}
\usepackage{tabularx}
\usepackage{makecell}
\usepackage{algorithmic,algorithm}
\usepackage{multicol}
\usepackage{multirow}
\usepackage{float}
\usepackage[T1]{fontenc}

\usepackage[utf8]{inputenc}

\usepackage{microtype}

\usepackage{inconsolata}

\usepackage{graphicx}

\title{Visual Program Distillation with Template-Based Augmentation}

\author{
Michal Shlapentokh-Rothman\qquad Yu-Xiong Wang\qquad Derek Hoiem\\
\normalsize {University of Illinois at Urbana-Champaign}\\ 
\normalsize 
\{\texttt{michal5,yxw,dhoiem}\}\texttt{@illinois.edu}}

\begin{document}
\maketitle
\begin{abstract}
Adapting visual programming  or prompting large language models (LLMs) to generate executable code for visual tasks like visual question answering (VQA) for \emph{specialized} tasks or domains remains challenging due to high annotation and inference costs. We propose a \emph{low-cost} visual program distillation method that can be used for models with at most 1 billion parameters and requires \emph{no} human-generated program annotations. We achieve this through synthetic data augmentation based on decoupling programs into higher-level skills, called \emph{templates}, and their corresponding arguments. Experimental results show that, with a relatively small amount of question/answer data, small language models can generate high-quality specialized visual programs with the added benefit of much faster inference.

\end{abstract}

\section{Introduction}

Visual programming~\cite{visprog,codevqa,vipergpt} refers to generating programs that invoke visual models to solve tasks such as answering questions about images, typically by prompting a very large language model (LLM) like GPT~\cite{gpt4} or Llama~\cite{llama}. 

\begin{figure}[t]
  \centering
  \resizebox{\columnwidth}{!}{%
    \includegraphics{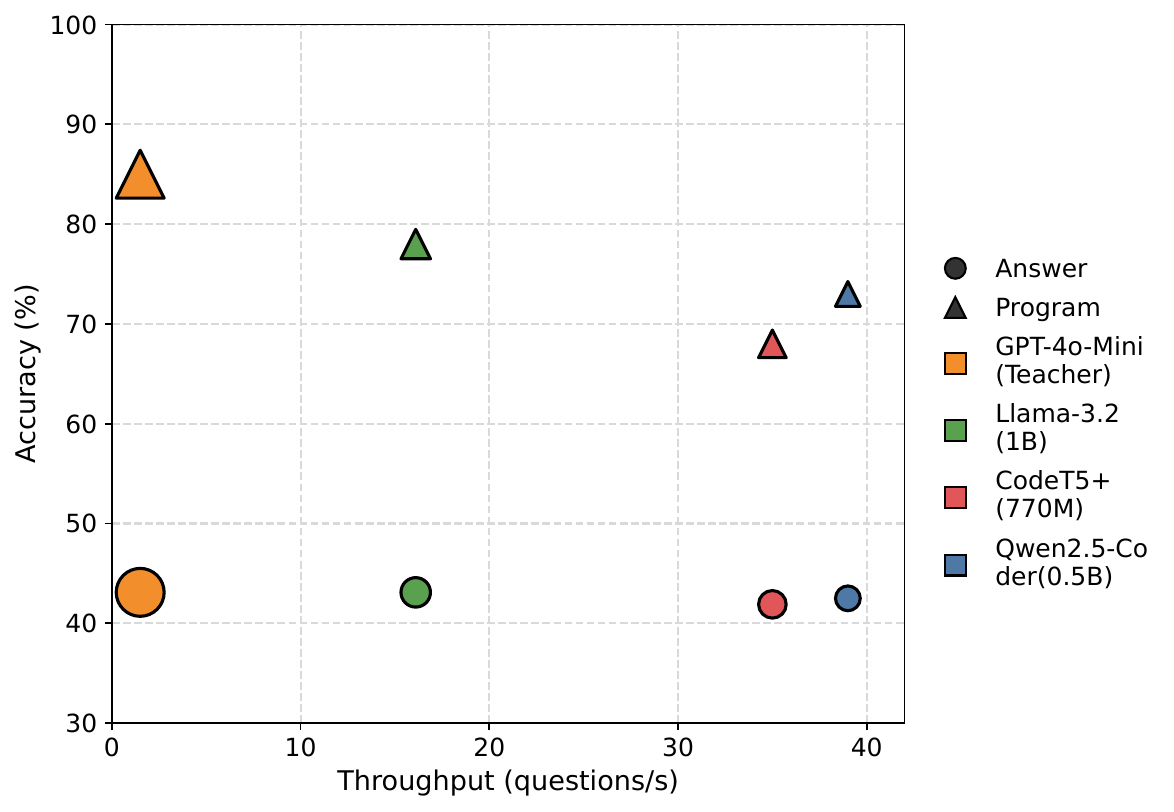}%
  }
  \caption{\textbf{Accuracy vs. Throughput for Visual Program Generation on GQA} Generalist LLMs (teacher models) offer high accuracy at the cost of low throughput and large model size (proportional to marker size). With our template-based augmentation method, specialized distilled student models achieve comparable performance on answer accuracy with a small percent of question/answer data ($\approx 0.1\%$) and no human program annotations.}
  \label{fig:teaser}
\end{figure}

Visual programming offers greater adaptability and customization for specialized applications such as a personal visual navigation assistant compared to a single vision-language model (VLM). In-context learning with proprietary or closed-source LLMs can generate correct visual programs for targeted applications but at the cost of long inference time and compute (see Figure~\ref{fig:teaser}). In addition, generating in-context examples (i.e. visual programs written by hand) requires a significant amount of human effort. Previous efforts~\cite{visrep} have made some progress in adapting smaller, open-source LLMs for dataset-specific visual programs but still suffer from high training and data costs. 

\textbf{Our goal is to create a \textcolor{Blue}{specialized} visual programming system with two key characteristics}:
\begin{enumerate}
    \item \textbf{Small Program Generator} Programs should be generated by models with $\leq1$ billion parameters, enabling fast inference and use on a wide variety of hardware. 
    \item \textbf{Limited Human Annotations} Minimize the number of required annotations, especially human-generated programs, for ease of adaptation.

\end{enumerate}

\begin{table*}[!htb]
\centering
\resizebox{\linewidth}{!}{ 
\begin{tabular}{lccccc}
\toprule
\textbf{Method} & \textbf{Task} & \textbf{Approach} & \textbf{Model Size} & \textbf{Train Size} & \textbf{Domain} \\ 
\midrule
VisRep~\cite{visrep} & Program Gen. & Self-Training & 7B & 10\% ($\approx$10K) & Visual Prog. \\
Distill Step-by-Step~\cite{distill-step-by-step} & Text Gen. & LLM Distillation & 220M-1B & 12.5\% & Q/A, Reasoning \\ 
CodePlan~\cite{code-plan} & Program Gen. & LLM Distillation & 770M & 100\% & General Coding \\
Template-Based Aug. (Ours) & Program Gen. & LLM Distillation & 500M-1B & 0.1\% ($\approx$1K) & Visual Prog. \\ 
\bottomrule
\end{tabular}
}
\caption{Comparison of approaches for program generation and LLM distillation. Our method enables LLM distillation for visual programming using only a small fraction of the training dataset and minimizing annotation costs.}
\label{tab:related}
\end{table*}

Our key insight in achieving such capabilities lies in decoupling the skill or procedure from the question-specific concept. We call the higher-level skills \textit{templates} and the concepts \textit{arguments}.  For example, the programs ``Count the red chairs" and ``Count the green bananas" have the same template,`find(arg1), verify property(arg2), count',  but different arguments: `red' (arg1) and `chairs' (arg2) vs `green' (arg1) and `bananas' (arg2) respectively. This decomposition facilitates creating synthetic examples by replacing arguments in the question and program, e.g. `Count the red apples' with corresponding argument substitution in the program. Given a small training dataset, we can use this process, which we refer to as \emph{template-based augmentation} to increase concept diversity. 

Combining template-based augmentation with recent advances in LLM distillation~\cite{distill-step-by-step}, we propose a low-cost visual program distillation method. Our approach requires \emph{no} human-generated example programs and uses only a small fraction of  question/answer pairs (at most $0.1\%$ of the training dataset). A teacher model leverages auto-context generation, where each generated program that produces the correct answer is added to the set of in-context examples. The resulting annotated programs form a dataset of question/program pairs that are augmented via the template-based method described above and used to train a small language model (SLM) that efficiently generates visual programs.



Our method uses a low-cost teacher model (e.g. GPT-Mini-4o) and costs only around \$1 per dataset. In addition, the specialized distilled models achieve much faster inference speed (up to $\approx 30.8$x) than the teacher models.

To validate our approach, we evaluate on widely adopted visual question answering (VQA) datasets and compare with using human-generated annotations, few-shot prompting and non-augmented distillation. Most existing work evaluates only on whether the generated answer is correct.  We find additional insights by also evaluating the correctness of the generated programs (as judged by a human) and student/teacher agreement, revealing substantial discrepancies among the different metrics.


We summarize our contributions as follows:
\begin{itemize}
    \item Introduce a template-based augmentation framework that distills large, generalist visual programming systems into small, specialized ones with minimal human effort. Template-based augmentation increases performance across all evaluated metrics.
    \item Empirically show auto-context generation achieves the same or better performance as human provided annotations when prompting a teacher model.
    \item Discuss and evaluate different metrics: answer accuracy, student/teacher agreement, and program accuracy for visual program evaluation. Surprisingly, we find the rate of error in answers is up to 3.8 times higher than that of the programs, suggesting that future visual programming work should focus on more effective APIs.

\end{itemize}

\section{Related Works}

We outline the most relevant related work below. A high-level comparison between our work and others can be found in Table~\ref{tab:related}.

\label{sec:related-works}
\paragraph{Visual Programming} A long line of work investigates generating and executing programs to perform visual tasks. Early approaches~\citep{neural_module_networks,learning_to_compose,learning_to_reason,johnson2017inferring}  generate programs and execute the programs with learned end-to-end neural modules. Based on the impressive code generation capabilities of LLMs~\citep{codegen_perf}, visual programming frameworks~\citep{vipergpt,visprog,codevqa} generate programs given an API and in-context examples and execute the programs with large pre-trained vision models. Visual programming has been applied to many different domains and applications including visual question answering, video question answering,  text-to-image generation, and robotics~\citep{t2img_vp,chameleon,video_vp,robotics_vp}. Unlike previous visual programming works, we only use small models ($\leq 1$ billion parameters) to generate programs.

One of the challenges that distinguishes visual programming from LLM tool use is the dependence of execution and evaluation on the output of underlying models. As noted in other works~\cite{visrep}, incorrect programs can return (after being executed) the correct answer even if the program is incorrect, which motivates our investigation into different evaluation metrics.  

 A concurrent work, Visual Unit Testing (ViUniT)~\cite{viunit}, generates visual unit tests for both visual program selection and self-training. ViUniT also evaluates the unit test quality and program correctness and reports similar differences (36\%) between program and answer accuracy.
 
\paragraph{Tool-Based Finetuning} Compared to prompt-based methods, there are relatively few methods focused on improving program generation in visual programming for specific tasks or in the general field of tool-based LLMs through finetuning. One of the main challenges is the lack of program annotations for input/output pairs. Language modeling has a long history designing self-supervised tasks, especially in pre-training~\citep{bert,t5}. Toolformer~\citep{toolformer}, uses a form of self-supervision, to create a training dataset to finetune an LLM on tool-use programs. For each question/answer pair in a pre-existing training dataset, an LLM is prompted to generate a corresponding program. If the program decreases the training loss (of the same LLM), the program annotation is added to a new dataset. Then the same LLM is finetuned on the generated dataset. A similar approach is used in Chain-of-Thought (CoT) finetuning~\citep{program_distillation},  program correction methods for specific datasets such as VDebugger~\citep{vdebugger} and specialized visual programming methods~\citep{visrep}, such as VisRep. LoRA~\citep{lora} is frequently used when finetuning LLMs on the generated data. 

In VisRep,~\citet{visrep} use a self-training approach similar to toolformer to finetunue an LLM for dataset-specific visual programming. Self-annotated programs are kept if the executed program returns the correct answer.  While there are some similarities between VisRep and our work, the underlying setting is different. VisRep focuses on improving program generation on an existing LLM (7B) through self-training for specialized tasks, while the goal of our work is to distill visual program generation from an LLM to a small model ($\leq1$B) for specialized tasks. Our work also does not require sampling across specific question types or correcting by hand different program annotations during training. We randomly sample the training set and do not write any program annotations or corrections.


\paragraph{Distillation}
Knowledge distillation~\citep{knowledgedistill,buciluǎ2006model}, where a large teacher model annotates unlabeled data that is then used to finetune a  smaller, weaker student model, is frequently used across many different applications including image recognition~\citep{patient_teacher}, masked language modeling~\citep{distillbert} and commonsense knowledge~\citep{commonsense_distill}. One of the difficulties of distillation is the need for a large number of unlabeled training examples which can be expensive to obtain. One way to compensate for this is to train the student model with a multi-task objective. The objective consists of a weighted sum of the cross-entropy on the original task and cross-entropy on a closely related part of the task such as the rationale in CoT~\citep{distill-step-by-step}.  CodePlan~\citep{code-plan} applies such an idea to code generation where the secondary objective is a natural language version or plan of the code, which has been shown to be effective in prompting-based work~\citep{self-plan-code}.

One downside to multi-objective prompting is that both the teacher and the student models have to generate additional output. In CodePlan, during inference, the model first generates the plan and then generates the desired code creating long inference time. Such steps might be necessary for complex code generation but are unnecessary for visual programming. Instead we use a relatively simple data augmentation method based on abstractions~\citep{craft} or higher level plans of existing programs which we refer to as templates. Templates do not require any additional forward passes to create and can be augmented using simple techniques such as word replacement~\cite{distill-augmentation}.

\paragraph{Automatic In-Context Example Generation} 
Prompt-based methods are more effective when in-context examples or demonstrations~\citep{cot} are included. In visual programming~\citep{vipergpt,visprog}, manually written in-context examples are used to adapt an API to a particular dataset. However, such an approach can be time consuming and not feasible for a large training set of input/output pairs. A common practice is to have an LLM self-annotate examples and keep the examples that produce the correct answer~\cite{auto-cot,codebison,webwise,selective-in-context}. Then during inference, retrieval augmented generation (RAG)~\cite{rag} can be performed to select the most relevant ones. We follow such an approach, referred to as auto-context generation by ~\citet{webwise} to generate in-context examples for the teacher model. 
\begin{figure*}
       \centering
    \includegraphics[width=0.95\linewidth]{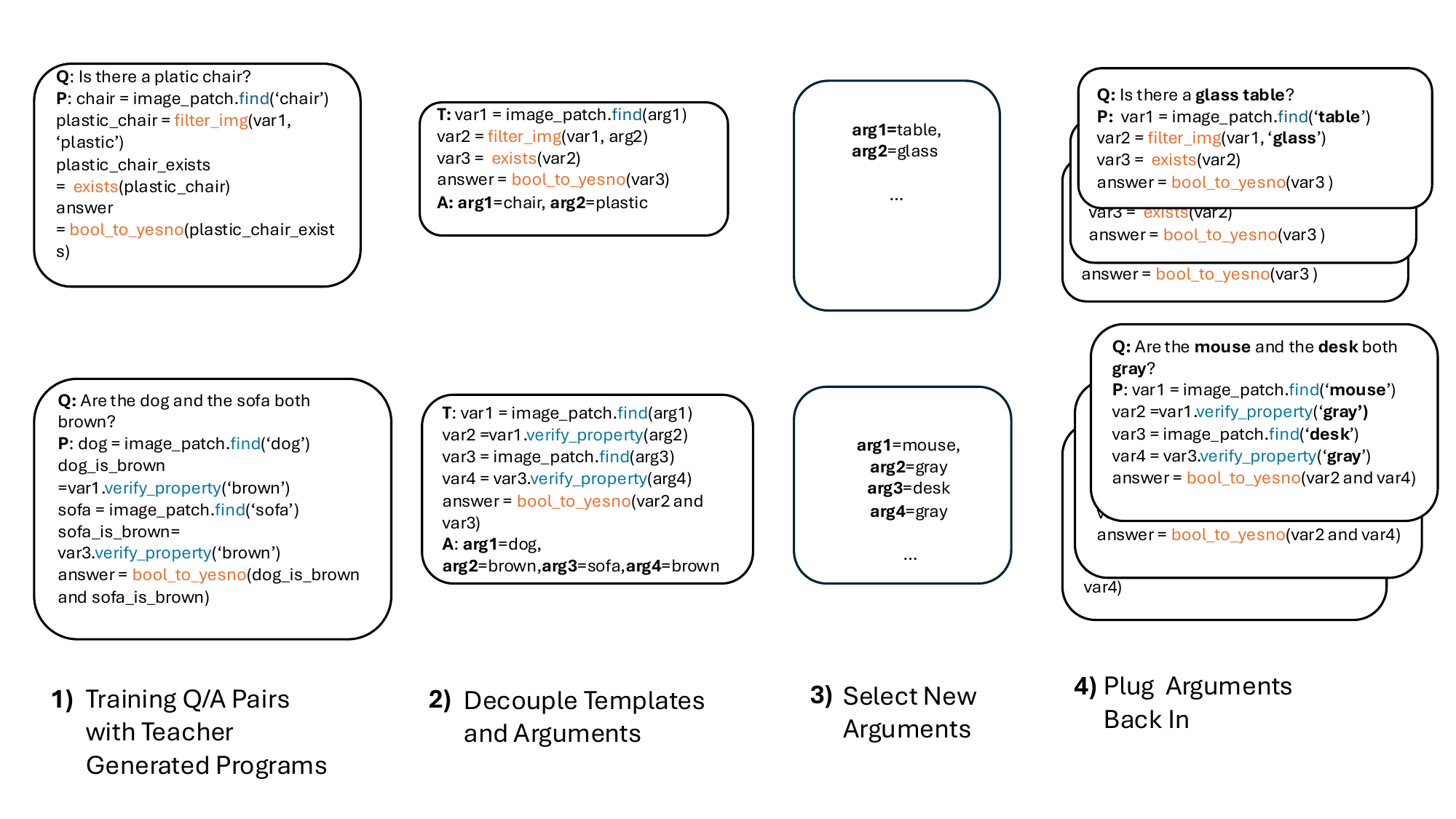}
    \caption{An overview of our augmentation method. Programs are first separated into templates and argument, new arguments are selected and plugged back into the question/program pair. Templates are created by renaming variables and removing question specific concepts. One single teacher generated question/program pair can turn into hundreds of new question/program pairs.  }
    \label{fig:template-main}
\end{figure*}

\section{Method}

\paragraph{Preliminaries} 
 Following the notation in ViperGPT~\cite{vipergpt}, the visual programming objective is to generate a program $z = \pi(q,p)$ with a program generator $\pi$, input query $q$ and prompt $p$ such that when executed with an execution engine $\phi$ and corresponding visual input $x$, $\phi(x,z)$ returns the correct answer. Prompt $p$ contains an API and dataset-specific in-context examples.

Given a teacher model, $\pi_t$, and a smaller student model ($\leq1$ billion parameters) $\pi_s$, our goal is to distill visual program generation from the teacher to the student for a specific dataset using a minimal number of answer annotations (question/answer pairs). We do not have any human provided program annotations (question/program/answer triplets), but we do know the API used for the teacher model.

There are three main steps to our approach: teacher annotation, data augmentation  and student training.

\subsection{Teacher Generated Program Annotation}
\label{sec:teach-gen}
We use GPT-4o-Mini~\cite{gpt40mini} as our teacher model since it is relatively cheap and has strong performance. Unlike traditional knowledge distillation, LLM based distillation requires the teacher (an LLM) to be given the appropriate context through in-context examples before starting annotation. Common practice is to have a human generate the examples.  We follow a process known as auto-context generation~\cite{webwise} to automatically generate such examples given answer annotations only. 


Auto-context generation follows a simple process similar to VisRep~\cite{visrep}:
\begin{enumerate}
    \item Teacher model predicts a program using API and in-context examples (if any). 
    \item Generated program is evaluated.
    \item If the answer returned by a generated program matches the ground truth answer, then the program is immediately added to the set of in-context examples. Otherwise the program is discarded.
\end{enumerate}

For efficiency, the in-context examples are sampled based on similarity to the question (i.e. RAG)~\cite{rag} once there are more than $x$ examples. We compute the cosine similarity with finetuned MP-Net-v2~\cite{sentence-bert} between an input and all in-context examples and select the $x=50$ highest scoring examples.

\lstset{
  language=Python,
  basicstyle=\ttfamily\footnotesize,
  columns=fullflexible,
  keepspaces=true,
  numbers=none,
  showspaces=false,
  showstringspaces=false,
  tabsize=4,
  breaklines=true,
}
\begin{table}[!htb]
    \centering
    \resizebox{\columnwidth}{!}{
    \begin{tabular}{p{4cm}p{8cm}}  
        \hline 
        \textbf{Template} &  
        \parbox{9cm}{
            \texttt{image\_patch = ImagePatch(image)}\\
            \texttt{var1 = image\_patch.find(`arg\_0')}\\
            \texttt{var2 = var2.classify(`arg\_1')}\\
            \texttt{var3 = image\_patch.find(`arg\_2')}\\
            \texttt{var4 = var3.classify(`arg\_3')}\\
            \texttt{answer = bool\_to\_yesno(var2 == var4)}
        }\\
    \hline 
       
    \makecell[l]{Are the \textcolor{red}{cat} and the \textcolor{blue}{tshirt} \\ the same \textcolor{Green4}{color}?} & 
    \makecell[l]{
    arg\_0 = \textcolor{red}{cat} \\  
    arg\_1 = \textcolor{Green4}{color}\\ 
    arg\_2 = \textcolor{blue}{tshirt} \\ 
    arg\_3 = \textcolor{Green4}{color}} \\ 
   
    \hline

    \makecell[l]{Is the \textcolor{red}{sofa} made from the \\ same \textcolor{Green4}{material} as the \textcolor{blue}{chair}?} & 
    \makecell[l]{
    arg\_0 = \textcolor{red}{sofa}\\
    arg\_1 = \textcolor{Green4}{material}\\
    arg\_2 = \textcolor{blue}{chair} \\  
    arg\_3 = \textcolor{Green4}{material}} \\ 
    \hline

    \makecell[l]{Is the \textcolor{red}{vase} the same \\ \textcolor{Green4}{shape} as the \textcolor{blue}{table}?} & 
    \makecell[l]{
    arg\_0 = \textcolor{red}{vase} \\
    arg\_1 = \textcolor{Green4}{shape} \\
    arg\_2 = \textcolor{blue}{table} \\
    arg\_3 = \textcolor{Green4}{shape}}\\
    \hline

    \end{tabular}
    }
    \caption{A template is a particular ordering of operations. The questions above all share the same template since they only differ in the arguments. We want to answer similar questions the same way and easily generate synthetic data.}
    \label{tab:template-question}
\end{table}





\subsection{Data Augmentation}
\label{sec:aug}
After the teacher generates a dataset of question/program pairs, our goal is to train a student model on these pairs only. At this point, corresponding answers and images of questions are not used.  Since the dataset is small, we use data augmentation to create a greater variety of question/program pairs. 

To understand the intuition behind our data augmentation method, consider the set of questions in ~\autoref{tab:template-question}. All of these questions compare properties of two objects. The general structure of each program is the same with the only difference coming from the inputs to the functions. If we answer one of these questions correctly and know that the remaining questions have the same structure, then all the remaining questions should also have that same structure or be consistent. An overview of our data augmentation method can be seen in Figure~\ref{fig:template-main}. We refer to this as "template-based augmentation." 

\lstset{
  language=Python,
  basicstyle=\ttfamily\footnotesize,
  columns=fullflexible,
  keepspaces=true,
  numbers=none,                    
  showspaces=false,
  showstringspaces=false,
  tabsize=4,
  breaklines=true,
}
\begin{table}[!ht]
    \centering
    \resizebox{\columnwidth}{!}{
   
        \begin{tabular}{p{4cm}p{9cm}}
        \hline 
        Example & Template \\
        \hline

        \parbox{4cm}{
          Is the \textcolor{blue}{blue} \textcolor{Green4}{car} the same \textcolor{red}{sha}\textcolor{Purple1}{pe} as the \textcolor{Gold3}{chair}?\\
          Is \textcolor{blue}{leather} \textcolor{Green4}{jacket} made of the same \textcolor{red}{mate}\textcolor{Purple1}{rial} as the \textcolor{Gold3}{shirt}?
        } & 
       
          \parbox{7cm}{
          \vspace{2pt}
            \texttt{image\_patch = ImagePatch(image)}\\
            \texttt{var1 = image\_patch.find(\textless \textcolor{Green4}{arg\_0}\textgreater)}\\
            \texttt{var2 = filter\_img(\textless \textcolor{blue}{arg\_1}\textgreater)}\\
            \texttt{var3 = var2.classify(\textless \textcolor{red}{arg\_2}\textgreater)}\\
            \texttt{var4 = image\_patch.find(\textless \textcolor{Gold3}{arg\_3}\textgreater)}\\
            \texttt{var5 = var4.classify(\textless \textcolor{Purple1}{arg\_4}\textgreater)}\\
           \texttt{answer =bool\_to\_yesno(var3 $==$ var5)}}\\
        \hline 
     
         \parbox{4cm}{
          What type of \textcolor{red}{food} is \textcolor{blue}{near} the \textcolor{Green4}{per}\textcolor{Gold3}{son}?\\
          What is the \textcolor{Green4}{veh}\textcolor{Gold3}{icle} \textcolor{blue}{next to} the \textcolor{red}{animal}?
        } & 
       
          \parbox{7cm}{
          \vspace{2pt}
            \texttt{image\_patch = ImagePatch(image)}\\
            \texttt{var1 = image\_patch.find(\textless \textcolor{Green4}{arg\_0}\textgreater)}\\
            \texttt{var2 = var1.crop\_position(\textless \textcolor{blue}{arg\_1}\textgreater)}\\
            \texttt{var3 = var2.find(\textless \textcolor{red}{arg\_2}\textgreater)}\\
            \texttt{answer $=$ var3.classify(\textless \textcolor{Gold3}{arg\_3}\textgreater)}}\\
        \hline
  
         \parbox{4cm}{
          Is the \textcolor{Green4}{car} to the \textcolor{red}{left or right} of the \textcolor{blue}{tree}?\\
         Is \textcolor{Green4}{pot} \textcolor{red}{above or below} the \textcolor{blue}{pan}?
        } & 
    
          \parbox{10cm}{
          \vspace{2pt}
            \texttt{image\_patch = ImagePatch(image)}\\
            \texttt{var1 = image\_patch.find(\textless \textcolor{Green4}{arg\_0}\textgreater)}\\
            \texttt{var2 = image\_patch.find(\textless \textcolor{blue}{arg\_1}\textgreater)}\\
            \texttt{answer $=$  choose\_relationship(var1,var2,\textless \textcolor{red}{arg\_2}\textgreater)}}\\
        \hline
        \end{tabular}
    }
    \caption{Some examples of questions and corresponding templates. Multi-colored words correspond to multiple arguments.}
    \label{tab:template-ex}
\end{table}

\paragraph{Templates}
We define a template as a specific ordering of functions, where a function is an API call to a visual model or python operation. Templates encode program structure independently of argument values. For example, if the program is

\begin{flushleft}
\texttt{image\_patch = ImagePatch(image)}\\
\texttt{dog = image\_patch.find(`dog')}\\
\texttt{answer = dog.classify(`color')}
\end{flushleft}
        
then the template would be

\begin{flushleft}
       \texttt{image\_patch = ImagePatch(image)}\\
        \texttt{var1 = image\_patch.find(\textless arg\textgreater)}\\
        \texttt{answer = var1.classify(\textless arg\textgreater)}\\

\end{flushleft}

Please see Table~\ref{tab:template-ex} for examples of questions and corresponding templates. 

Templates can be considered high-level plans used for plan-based distillation~\cite{code-plan}. The main advantage of templates is that they can be extracted directly from the program. There is no need to generate extra output or perform multiple forward passes for a single question. 
 \begin{figure*}[!htb]

    \centering
    \includegraphics[width=1\linewidth]{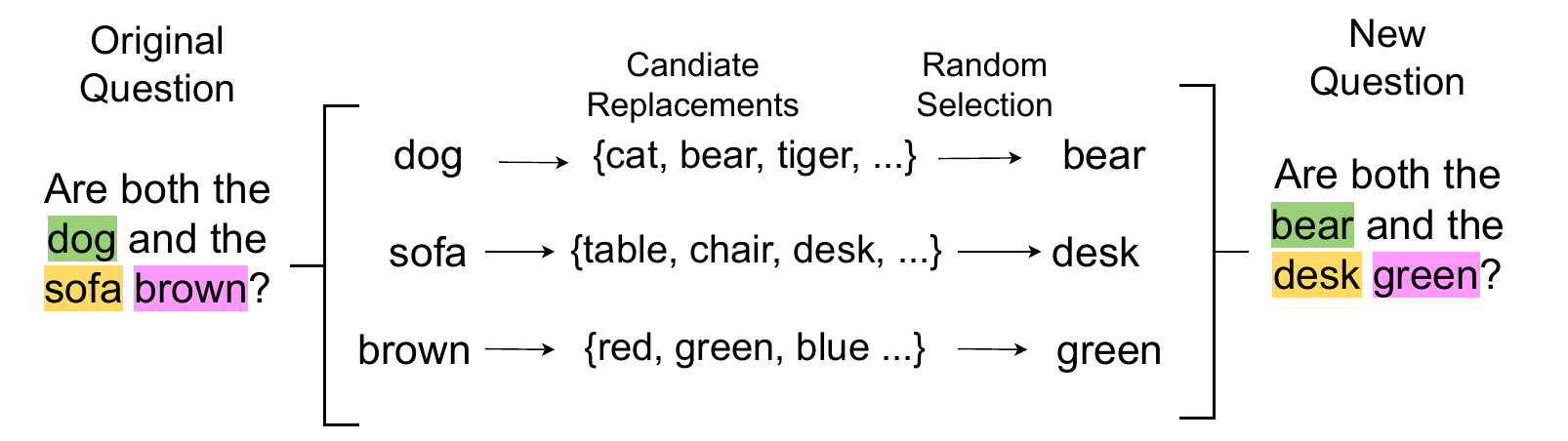}
    \caption{An example of our data augmentation approach. Both the new and old question have the same template, so the template matcher output should predict the same template for both. The arguments for the new and old programs are different. But, in the arguments,  (dog, sofa, brown) should be replaced with (bear, desk, green).  }
    \label{fig:sys-diag}
\end{figure*}

\paragraph{Template Extraction} Given the program annotations from the teacher model, we extract templates and corresponding arguments from each program similar to the abstraction method of CRAFT~\cite{craft}. Extracting is quite simple: replace specific variable names with generic ones and put in placeholders for each argument. The variable renaming and extraction can be done in seconds with abstract syntax trees and regular expression matching. Note that while the template \emph{extraction} algorithm is deterministic, the actual templates are determined solely by the annotations from the teacher model.  The code can be found in Appendix~\ref{sec:rename}.  

\paragraph{Augmentation} We can use the decomposition of programs into templates and arguments to generate synthetic data similar to how masked language modeling is used in BERT~\cite{bert}. As can be seen in Table~\ref{tab:template-question}, for many questions and programs, the arguments appear directly in the question. Consider the example in Figure~\ref{fig:sys-diag}: "Are both the dog and the sofa brown?" The arguments are dog, sofa, and brown. Once we find similar words for each, we can simply replace them in the sentence. At this point,  we also have the program, since the template and arguments are known.  

There are two different methods for generating word replacements. For the GQA~\citep{gqa} dataset, the possible word replacements come from ViperGPT~\cite{vipergpt} except for two special circumstances. The first is when arguments are not in the question. For example, the question "What is that made from?" could have a program where the arguments are "object" (with the find function) and "material" (with the classify function). The second is when the program consists of only calling a VLM and the input is the entire question. For the special cases, we use GPT~\cite{gpt40mini} to generate similar questions. 
For VQAv2~\cite{vqa_v2}, we use BERT~\cite{bert} when an argument is a single word and BART~\cite{bart} to replace phrases. If possible, we replace arguments with arguments of the same type, e.g. an attribute is replaced by an attribute. 

 During training, each argument in each question has a 50\% chance of being replaced. If an argument is to be replaced, we then uniformly sample among the possible replacements. In the GQA dataset, for some arguments, like objects, the number of possible replacements is quite large (e.g. greater than 1500), while for arguments like directions such as left, right, etc. the number of replacements is small (e.g. fewer than 10). Some examples of categories and more details on word replacement can be found in Appendix~\ref{sec:replace}.

\paragraph{Student Training}
 Given the augmented dataset, we perform LoRA-based finetuning on the student model. The input to the student model is the question and the output is the visual program. Since the augmentation method produces both questions and corresponding programs, we use a next-token prediction loss for training. 

\section{Experiments}
\label{sec:exp}



\paragraph{Experimental Setup}
For all experiments, the teacher model is GPT-4o-mini. Several student models are trained with LoRA~\cite{lora}: Qwen2.5-Coder-0.5B~\cite{qwen25coder}, CodeT5~\cite{codet5plus} and Llama-3.2-1B~\cite{llama3}. \hspace{-1em}We use the GQA dataset~\cite{gqa} and VQAv2~\cite{vqa_v2} datasets for our experiments. All student models are trained and evaluated on a single dataset unless indicated by the word `joint'. We evaluate on the full GQA test-dev split and randomly sample 10,000 questions from the VQAv2 validation split. GQA is evaluated using exact match and VQAv2 is evaluated based on annotator/answer agreement as in the original benchmark. Experiments on additional datasets can be found in Appendix~\ref{sec:add_results}.
\begin{table}[ht]
    \centering
    \resizebox{\columnwidth}{!}{
    \begin{tabular}{lcc|c}
        \toprule
        \textbf{Dataset} & \textbf{Human Generated} & \textbf{LLM Generated} & \textbf{Performance} \\
        \midrule
        \multirow{4}{*}{\textbf{GQA}} 
            & 0  & 0   & 35.1 \\
            & 25 & 0   & 38.5 \\ 
            & 0  & 474 (out of 1000) & \textbf{43.1} \\
            & 25 & 510 (out of 1000) & \textbf{43.1} \\ 
        \midrule
        \multirow{4}{*}{\textbf{VQAv2}} 
            & 0  & 0   & 47.4 \\ 
            & 25 & 0   & 54.2 \\ 
            & 0  & 286 (out of 500) & \textbf{60.3} \\
            & 25 & 283 (out of 500) & \textbf{57.8} \\
        \bottomrule
    \end{tabular}}
    \caption{Teacher performance compared across varying numbers of human-generated and LLM generated in-context examples (i.e. auto-context generation). Auto-Context Generation achieves the same or better performance than human-generated program annotations.}
    \label{tab:acg}
\end{table}

\begin{table*}[!ht]
  \centering
  \scriptsize
  \setlength{\tabcolsep}{4pt}         
  \renewcommand{\arraystretch}{0.9}   
  \begin{tabularx}{\linewidth}{l X X}
    \toprule
    \textbf{Error} & \textbf{Example} & \textbf{Explanation} \\
    \midrule
    Not executable &
      \textbf{Q:} Is the chair left or right? 
      \newline 
      \makecell[l]{\textbf{P:}
        \texttt{side = choose\_relationship}\\
        \texttt{(chair, img, [`left',`right'])}%
      }
      &
      choose\_relationship requires a list as input \\
    \addlinespace[2pt]  
    API Violation &
      \textbf{Q:} What color is the running dog on the left?  
      \newline
      \textbf{P:} \texttt{crop\_position
      (`running left',img)}
      &
      “running left” isn’t a valid direction/preposition. \\
    Contradicts Question &
      \textbf{Q:} What color is the car above the road?  
      \newline 
      \textbf{P:} \texttt{below\_road = crop\_position(`below', road)}
      &
      The question states that the car is above the road, not below. \\
    Does Not Answer Question &
      \textbf{Q:} Are there two tables?
      \newline 
      \textbf{P:} \texttt{num = count(image.find(`table'))}
      &
      The question asks if there are two tables, not how many tables there are. \\
    Does Not Include All Question Information &
      \textbf{Q:} Is the blue toy small?  
      \newline 
      \textbf{P:} \texttt{toy.verify\_property(`small')}
      &
      The question specifically asks about the blue toy. \\
    \bottomrule
  \end{tabularx}
  \caption{Common errors encountered during qualitative evaluation.}
  \label{tab:error_classification}
\end{table*}

We use a slightly modified API from the original ViperGPT paper ~\cite{vipergpt}. The main differences are some additional functions (to reduce program length) and removal of the use of a VLM if earlier parts of a program fail. Please see Appendix~\ref{sec:viperapi} for more details. For visual models, we use InstructBLIP (Flan-T5 XL)~\cite{instructblip} for general visual queries, Owl-ViT2~\cite{owlvit2} for detection and CLIP~\cite{clip} for classification.
\begin{table*}[!htb]
  \centering
  \scriptsize
  \setlength{\tabcolsep}{4pt}          
  \renewcommand{\arraystretch}{0.9}    
  \resizebox{\linewidth}{!}{%
    \begin{tabular}{l rrr rrr}
      \toprule
      Method
        & \multicolumn{3}{c}{\textbf{GQA}}
        & \multicolumn{3}{c}{\textbf{VQA}} \\
      \cmidrule(lr){2-4} \cmidrule(lr){5-7}
        & Answer Acc & Prog.\ Acc & Stu/Tea Agree.
        & Answer Acc & Prog.\ Acc & Stu/Tea Agree. \\
      \midrule
      \multicolumn{7}{l}{\textbf{Few-shot prompting}} \\
      Llama\,3.2-1B   & 19.0 & 22 & –    & 23.3 & 40 & –    \\
      Qwen Coder\,2.5-0.5B  & 18.1 & 14 & – & 25.0 & 16  & – \\
      \midrule
      \multicolumn{7}{l}{\textbf{Teacher}} \\
      GPT-4o-Mini & 43.1 & 85 & – & 60.3 & 93 & – \\
      \midrule
      \multicolumn{7}{l}{\textbf{Distilled Students w/Aug.}} \\
      Qwen Coder\,2.5-0.5B (single) & 42.5 & 71 & 78.8 & 60.8 & 73  & 74.2 \\
      Qwen Coder\,2.5-0.5B (joint)  & 
      \textbf{43.1} & \textbf{78} & 78.2 & 60.3 & 79  & 70.5 \\
      Code-T5-770M (single)          & 41.9 & 68 & 64.5 & \textbf{61.1} & 80 & \textbf{76.1} \\
      Llama\,3.2-1B (single)            & \textbf{43.1} & 73 & \textbf{81.2} & 60.3 & \textbf{82}  & 73.6 \\
      \bottomrule
    \end{tabular}%
  }
  \caption{Comparison of few-shot prompting and student models trained w/ augmented distillation across different metrics. Few-shot open, source models have poor performance but template-augmented student distillation has comparable performance to the teacher model, GPT-4o-Mini.}
  \label{tab:compare-results}
\end{table*}

\paragraph{Auto-Context Generation}  Before training the student model, we adapt the teacher to the VQA domain. We investigate the effectiveness of using human-generated programs and programs generated by the teacher model validated by answer correctness. As our aim is to enable visual programming with low cost and effort, we compare using a small number (25) of human-generated programs to generating 1,000 programs (yielding hundreds of validated programs).

As shown in Table~\ref{tab:acg}, human-generated programs provide significant improvement vs. no examples, but provide no further benefit given self-generated examples. This finding supports using self-generation to improve the teacher model, as question-answer pairs require much less expertise and time to provide.

\paragraph{Evaluation Metrics}

All approaches are evaluated with two metrics: answer accuracy and program accuracy. Student models are also evaluated on student/teacher answer agreement or if student and teacher programs return the same answer, regardless of correctness. Program accuracy involves a human (the authors) manually evaluating each program (without execution) for correctness. We randomly sample 100 questions from each dataset for program correctness evaluation. The same 100 questions are used to evaluate each method. To ensure the fairness of our program evaluation, we enlist a second annotator to evaluate program correctness on 50 examples: 25 correct and 25 incorrect. The second annotator agreed with the original program evaluation on 47 out of 50 of the examples. Program annotations have been released so others can contribute.\footnote{\url{https://github.com/michalsr/vp_distill_templates/tree/main/program_correctness_annotations}}

There are multiple ways to determine if a program is correct. We generally assume a program is correct unless it violates one of these criteria:

- \textbf{Not Executable} The program must be executable and return the correct data type (a string for VQA datasets). 

- \textbf{API Violation} Visual programming APIs are designed to follow basic visual knowledge and reasoning. For example, the `find' function is used with nouns while `verify\_property' is generally used for attributes. Clear violations such as trying to find an attribute (e.g. `find(green)') or cropping with a verb (e.g. `crop\_position(running)') are considered incorrect. 

- \textbf{Contradicts Question} Programs that assume a statement that directly conflicts with a statement in the question. For example, assuming an object is on the left, when the question states it is on the right. 

- \textbf{Does Not Answer Question}  Programs that do not answer the question, even if the program correctly follows the API, are incorrect. Common examples are returning yes/no instead of choosing between two options such as left or right.

- \textbf{Does Not Include Vital Information From the Question} If the question includes details about an object, then those details must be in the program. 

Some examples can be seen in Table~\ref{tab:error_classification}.

\paragraph{Few-Shot Comparison} First, we compare our distilled student models with the most closely related setting: prompt-based visual program generation. Other works, such as VisRep~\cite{visrep}, have different objectives and use carefully curated and larger datasets, and thus are not included in our comparisons. Furthermore, VisRep evaluates on selected subsets of various datasets, and the specific subsets, models, and code have not been made publicly available for comparison.

Two models, Llama-3.2-1B and Qwen2.5-Coder-0.5B, are evaluated using both few-shot prompting and template-based distillation. CodeT5 is also evaluated with template-based distillation. There are two versions of distilled Qwen2.5-Coder-0.5B: one version is trained (and evaluated) on GQA and VQA separately (denoted by `single') and the other is trained on \textbf{both} datasets (denoted by `joint').  

The same 25 human generated in-context examples (incorporated into the ViperGPT~\cite{vipergpt} API) used in the Auto-Context Generation experiments above are used per dataset for each few-shot model.  The distilled models are trained with 0.1\% of the data (474 question/answer pairs for GQA and 286 question/answer pairs for VQAv2). All program annotations come from the teacher model. 

\begin{table*}[!htb]
  \centering
  \scriptsize
  \renewcommand{\arraystretch}{0.8}
  \resizebox{\linewidth}{!}{%
    \begin{tabular}{l rrr rrr}
      \toprule
      Method
        & \multicolumn{3}{c}{\textbf{GQA}}
        & \multicolumn{3}{c}{\textbf{VQA}} \\
      \cmidrule(lr){2-4} \cmidrule(lr){5-7}
        & Answer Acc & Prog.\ Acc & Stu/Tea Agree.
        & Answer Acc & Prog.\ Acc & Stu/Tea Agree. \\
      \midrule
        Qwen Coder\,2.5-0.5B w/o Aug (single)
        & 41.8 & 69 & \textbf{73.0} & \textbf{60.2} & 75 & 72.2 \\
      Qwen Coder\,2.5-0.5B w/ Aug (single)
        & (+\textcolor{Green4}{{0.7}})42.5 & (+\textcolor{Green4}{{2}})71  & (+\textcolor{Green4}{{5.8}})\textbf{78.8} & (+\textcolor{Green4}{{0.6}})\textbf{60.8}  & (-\textcolor{Red}{{2}})73 & (+\textcolor{Green4}{{2.0}})74.2   \\
 
      \midrule
    
      Qwen Coder\,2.5-0.5B w/o Aug (joint)
        & \textbf{41.7} & 69 & \textbf{73.5} & 58.8 & 72 & \textbf{67.6} \\
        Qwen Coder\,2.5-0.5B w/ Aug (joint)
        & (+\textcolor{Green4}{{1.4}}) \textbf{43.1}& (+\textcolor{Green4}{{9}})78  & (+\textcolor{Green4}{{4.7}})\textbf{78.2}  & (+\textcolor{Green4}{{1.5}})60.3  & (+\textcolor{Green4}{{6}})78 & (+\textcolor{Green4}{{2.9}})\textbf{70.5}  \\
      \midrule
        Code-T5-770M w/o Aug (single)
        & 41.1 & \textbf{48} & \textbf{59.8} & 60.8 & 79 & \textbf{72.0} \\
      Code-T5-770M w/ Aug (single)
        & (+\textcolor{Green4}{{0.8}})41.9  & (+\textcolor{Green4}{{20}})\textbf{68} & (+\textcolor{Green4}{{4.7}})\textbf{64.5}  & (+\textcolor{Green4}{{0.3}})61.1 & (+\textcolor{Green4}{{1}})80 & (+\textcolor{Green4}{{4.1}})\textbf{76.1} \\

      \midrule
        Llama\,3.2-1B w/o Aug (single)
        & \textbf{40.0} & \textbf{61} & \textbf{70.1} & 60.2 & 79 & 72.6 \\
      Llama\,3.2-1B w/ Aug (single)
        & (+\textcolor{Green4}{{3.1}})\textbf{43.1}   & (+\textcolor{Green4}{{12}})\textbf{73} & (+\textcolor{Green4}{{11.1}})\textbf{81.2}  & (+\textcolor{Green4}{{0.1}})60.3  & (+\textcolor{Green4}{{3}})82 & (+\textcolor{Green4}{{1}})73.6  \\

      \bottomrule
    \end{tabular}%
  }
  \renewcommand{\arraystretch}{1.0}
  \caption{Effect of augmentation. On average, template-based augmentation improves performance on nearly all models, particularly on student/teacher agreement and program accuracy. \textbf{Bold} results are statistically significant.}
  \label{tab:data-aug}
\end{table*}


\textbf{Few-Shot Results} We can draw several observations from the results in Table~\ref{tab:compare-results}. Comparing different methods, we see that few-shot prompting on small open-source models results in extremely low performance but with a small amount of data and a strong enough teacher, the same models achieve similar performance to the auto-context teacher, especially on answer accuracy. Program accuracy improves significantly from distillation but a gap still remains indicating room for improvement. 

The results across different metrics are a bit more surprising. On all of the student models, the average difference between student/teacher answer agreement and answer accuracy is \textbf{33\%} for GQA and \textbf{13\%} for VQAv2, indicating that answer accuracy underestimates distillation performance. The noisiness of answer accuracy is illustrated even more by program accuracy performance.

The program accuracy results are the most surprising. For all of the methods, program accuracy is higher than answer accuracy especially for the teacher model GPT-4o-Mini  which has 85\% vs. 43\% and 93\% vs. 60\% program vs answer accuracy for GQA and VQAv2 (respectively). The significant difference among the metrics indicates that most errors in visual programming systems with proprietary LLMs are \emph{not from the programs} but from the API and visual models. An additional analysis of program errors is in Appendix~\ref{sec:program-analysis}.


\paragraph{Data Augmentation} Next, we ablate the effect of data augmentation on distillation. For each model, we train with and without distillation on both datasets and measure performance across the three metrics. From the results in Table~\ref{tab:data-aug}, we see that data augmentation improves performance across all metrics for nearly all models and datasets. Augmentation has more of an effect on GQA compared to VQAv2. Both datasets have fairly easy questions, but VQAv2 has many questions that are correctly answered by either calling a vision-language model (the simple\_query) function or simple counting, which involves two functions (find and count). The relative increase is more notable for program accuracy (average 6.4) and student/teacher agreement (average 4.5) compared to answer accuracy (average 1.1), again indicating that answer accuracy does not fully capture model behavior.

\paragraph{Cost and Efficiency} One of our objectives is low annotation costs and fast inference time.  As shown in Table~\ref{tab:efficiency}, the total annotation cost for program auto-generation on both datasets is less than a dollar. GQA costs a bit more because of the larger size and during augmentation, GPT-4o-Mini is called when arguments do not appear in the question. By using a small model, inference speed also greatly increases. 
\begin{table}[h]
\centering
\resizebox{\columnwidth}{!}{%
\begin{tabular}{@{}lrr@{}}
\toprule
\textbf{Metric} & \textbf{GQA} & \textbf{VQAv2} \\
\midrule
\multicolumn{3}{@{}l}{\textbf{Teacher}} \\
Program Generation Cost (Total) & \$0.69 & \$0.26 \\
GPT-4o-Mini Inference Time (q/s) & 1.3 & 1.3 \\
\midrule 
\multicolumn{3}{@{}l}{\textbf{Distilled Students}} \\
Qwen2.5-Coder Inference Time (q/s) & 39.2 & 40.4 \\
CodeT5 Inference Time (q/s) & 35.0 & 32.7 \\
Llama-3.2-1B Inference Time (q/s) & 16.1 & 18.6 \\
\bottomrule
\end{tabular}%
}
\caption{Annotation cost and inference time for student and teacher models. Student inference time is much faster than teacher inference time.}
\label{tab:efficiency}
\end{table}

\section{Conclusion}

 Our experiments demonstrate models trained with template-based visual program distillation can become specialized and efficient visual program generators at a small cost.  Auto-context generation removes the burden of human generated program annotations while still retaining the same performance. The results also show how commonly used metrics for visual programming, do not fully capture the performance. Human program verification reveals that on the best models, programs are likely not the source of errors and that future work should focus on the API and visual models, not program generation. For less than \$6, (\$5 for student training on the cloud and \$1 for annotation), a 500-M coding model can become a visual program generator. We anticipate that the use of template-based visual program distillation will enable users and researchers to iterate more quickly on various visual programming systems and broaden their use for targeted applications.

\section{Limitations and Future Work}
There are several limitations and areas for future work:

\textbf{Specialization}. Our method distills a specialized VQA model that is not intended to provide the same breadth of capability as the original LLM.  Our specialized models, however, are much faster and cheaper for inference and easy to produce, making them suitable for targeted application.

\textbf{Effort Required for Program Accuracy}. While program accuracy is an important metric, it also requires a significant amount of human effort, making it difficult to evaluate on a large scale. A promising area for future work includes automatic evaluation of program accuracy.

\textbf{Reliance on teacher model and API}. Our method relies on the quality of the teacher model, which in turn is dependent on the quality of the API and general prompt. A future challenge is how to learn from a weaker teacher and/or unreliable API. 

\textbf{Teacher Data Efficiency}. Following prior works, if the answer returned by a program was incorrect, it was not used. However, LLMs have the ability to self-correct given the appropriate feedback. In settings where question/answer pairs are limited, such an approach could be more cost-effective than discarding the examples. Areas for future work include incorporating self-correction methods or using program correction models such as VDebugger~\cite{vdebugger}. 

\textbf{Program Execution Time}. We evaluate the time to generate programs, but the time to execute with API calls to several visual models, can be much greater (3.4s / query in our implementation with high variance) and requires significant engineering effort to make efficient.


\textbf{Limited Program Complexity}. Existing VQA datasets are relatively simple and most work on visual programming is limited to tasks where the programs can be generated in a single step by an LLM. Most real world applications are multi-step and would require more complex reasoning and knowledge skills than in the evaluation datasets. 

\section{Societal Implications} Distilled student models inherit existing biases from LLMs so care is required when deploying the models in production. There are also privacy risks, so safeguards should be taken to prevent data leakage. 

\paragraph{Acknowledgement}
 This research is supported in part by NSF Grant 2106825, NIFA Award 2020-67021-32799, ONR N00014-23-1-2383 and ONR N00014-21-1-2705. This work used the Delta system at the National Center for Supercomputing Applications through allocation CIS230398 from the Advanced Cyberinfrastructure Coordination Ecosystem: Services \& Support (ACCESS) program~\cite{access}, which is supported by National Science Foundation grants \#2138259, \#2138286, \#2138307, \#2137603, and \#2138296.
\bibliography{egbib}

\begin{thebibliography}{54}
\providecommand{\natexlab}[1]{#1}

\bibitem[{Achiam et~al.(2023)Achiam, Adler, Agarwal, Ahmad, Akkaya, Aleman, Almeida, Altenschmidt, Altman, Anadkat et~al.}]{gpt4}
Josh Achiam, Steven Adler, Sandhini Agarwal, Lama Ahmad, Ilge Akkaya, Florencia~Leoni Aleman, Diogo Almeida, Janko Altenschmidt, Sam Altman, Shyamal Anadkat, et~al. 2023.
\newblock Gpt-4 technical report.
\newblock \emph{arXiv preprint arXiv:2303.08774}.

\bibitem[{Andreas et~al.(2015)Andreas, Rohrbach, Darrell, and Klein}]{neural_module_networks}
Jacob Andreas, Marcus Rohrbach, Trevor Darrell, and Dan Klein. 2015.
\newblock \href {https://api.semanticscholar.org/CorpusID:5276660} {Neural module networks}.
\newblock \emph{2016 IEEE Conference on Computer Vision and Pattern Recognition (CVPR)}, pages 39--48.

\bibitem[{Andreas et~al.(2016)Andreas, Rohrbach, Darrell, and Klein}]{learning_to_compose}
Jacob Andreas, Marcus Rohrbach, Trevor Darrell, and Dan Klein. 2016.
\newblock \href {https://doi.org/10.18653/v1/N16-1181} {Learning to compose neural networks for question answering}.
\newblock In \emph{Proceedings of the 2016 Conference of the North {A}merican Chapter of the Association for Computational Linguistics: Human Language Technologies}, pages 1545--1554, San Diego, California. Association for Computational Linguistics.

\bibitem[{Bareiss et~al.(2022)Bareiss, Souza, d’Amorim, and Pradel}]{codegen_perf}
Patrick Bareiss, Beatriz Souza, Marcelo d’Amorim, and Michael Pradel. 2022.
\newblock \href {https://api.semanticscholar.org/CorpusID:249375385} {Code generation tools (almost) for free? a study of few-shot, pre-trained language models on code}.
\newblock \emph{ArXiv}, abs/2206.01335.

\bibitem[{Beyer et~al.(2021)Beyer, Zhai, Royer, Markeeva, Anil, and Kolesnikov}]{patient_teacher}
Lucas Beyer, Xiaohua Zhai, Am{\'e}lie Royer, Larisa Markeeva, Rohan Anil, and Alexander Kolesnikov. 2021.
\newblock \href {https://api.semanticscholar.org/CorpusID:235376877} {Knowledge distillation: A good teacher is patient and consistent}.
\newblock \emph{2022 IEEE/CVF Conference on Computer Vision and Pattern Recognition (CVPR)}, pages 10915--10924.

\bibitem[{Bird et~al.(2009)Bird, Klein, and Loper}]{nltk}
Steven Bird, Ewan Klein, and Edward Loper. 2009.
\newblock \emph{{Natural Language Processing with Python}}.
\newblock O'Reilly Media.

\bibitem[{Boerner et~al.(2023)Boerner, Deems, Furlani, Knuth, and Towns}]{access}
Timothy~J. Boerner, Stephen Deems, Thomas~R. Furlani, Shelley~L. Knuth, and John Towns. 2023.
\newblock \href {https://doi.org/10.1145/3569951.3597559} {Access: Advancing innovation: Nsf’s advanced cyberinfrastructure coordination ecosystem: Services \& support}.
\newblock In \emph{Practice and Experience in Advanced Research Computing 2023: Computing for the Common Good}, PEARC '23, page 173–176, New York, NY, USA. Association for Computing Machinery.

\bibitem[{Buciluǎ et~al.(2006)Buciluǎ, Caruana, and Niculescu-Mizil}]{buciluǎ2006model}
Cristian Buciluǎ, Rich Caruana, and Alexandru Niculescu-Mizil. 2006.
\newblock Model compression.
\newblock In \emph{Proceedings of the 12th ACM SIGKDD international conference on Knowledge discovery and data mining}, pages 535--541.

\bibitem[{Cho et~al.(2024)Cho, Zala, and Bansal}]{t2img_vp}
Jaemin Cho, Abhay Zala, and Mohit Bansal. 2024.
\newblock Visual programming for step-by-step text-to-image generation and evaluation.
\newblock \emph{Advances in Neural Information Processing Systems}, 36.

\bibitem[{Dai et~al.(2023)Dai, Li, Li, Tiong, Zhao, Wang, Li, Fung, and Hoi}]{instructblip}
Wenliang Dai, Junnan Li, Dongxu Li, Anthony Tiong, Junqi Zhao, Weisheng Wang, Boyang Li, Pascale Fung, and Steven Hoi. 2023.
\newblock \href {https://openreview.net/forum?id=vvoWPYqZJA} {Instruct{BLIP}: Towards general-purpose vision-language models with instruction tuning}.
\newblock In \emph{Thirty-seventh Conference on Neural Information Processing Systems}.

\bibitem[{Devlin et~al.(2019)Devlin, Chang, Lee, and Toutanova}]{bert}
Jacob Devlin, Ming-Wei Chang, Kenton Lee, and Kristina Toutanova. 2019.
\newblock \href {https://api.semanticscholar.org/CorpusID:52967399} {Bert: Pre-training of deep bidirectional transformers for language understanding}.
\newblock In \emph{North American Chapter of the Association for Computational Linguistics}.

\bibitem[{Dubey et~al.(2024)Dubey, Jauhri, Pandey, Kadian, Al-Dahle, Letman, Mathur, Schelten, Yang, Fan et~al.}]{llama3}
Abhimanyu Dubey, Abhinav Jauhri, Abhinav Pandey, Abhishek Kadian, Ahmad Al-Dahle, Aiesha Letman, Akhil Mathur, Alan Schelten, Amy Yang, Angela Fan, et~al. 2024.
\newblock The llama 3 herd of models.
\newblock \emph{arXiv preprint arXiv:2407.21783}.

\bibitem[{Goyal et~al.(2017)Goyal, Khot, Summers{-}Stay, Batra, and Parikh}]{vqa_v2}
Yash Goyal, Tejas Khot, Douglas Summers{-}Stay, Dhruv Batra, and Devi Parikh. 2017.
\newblock Making the {V} in {VQA} matter: Elevating the role of image understanding in {V}isual {Q}uestion {A}nswering.
\newblock In \emph{Conference on Computer Vision and Pattern Recognition (CVPR)}.

\bibitem[{Gupta and Kembhavi(2023)}]{visprog}
Tanmay Gupta and Aniruddha Kembhavi. 2023.
\newblock Visual programming: Compositional visual reasoning without training.
\newblock In \emph{Proceedings of the IEEE/CVF Conference on Computer Vision and Pattern Recognition}, pages 14953--14962.

\bibitem[{Gurari et~al.(2018)Gurari, Li, Stangl, Guo, Lin, Grauman, Luo, and Bigham}]{vizwiz}
Danna Gurari, Qing Li, Abigale~J Stangl, Anhong Guo, Chi Lin, Kristen Grauman, Jiebo Luo, and Jeffrey~P Bigham. 2018.
\newblock Vizwiz grand challenge: Answering visual questions from blind people.
\newblock In \emph{CVPR}.

\bibitem[{Hinton et~al.(2015)Hinton, Vinyals, and Dean}]{knowledgedistill}
Geoffrey~E. Hinton, Oriol Vinyals, and Jeffrey Dean. 2015.
\newblock \href {https://api.semanticscholar.org/CorpusID:7200347} {Distilling the knowledge in a neural network}.
\newblock \emph{ArXiv}, abs/1503.02531.

\bibitem[{Hsieh et~al.(2023)Hsieh, Li, Yeh, Nakhost, Fujii, Ratner, Krishna, Lee, and Pfister}]{distill-step-by-step}
Cheng-Yu Hsieh, Chun-Liang Li, Chih-kuan Yeh, Hootan Nakhost, Yasuhisa Fujii, Alex Ratner, Ranjay Krishna, Chen-Yu Lee, and Tomas Pfister. 2023.
\newblock \href {https://doi.org/10.18653/v1/2023.findings-acl.507} {Distilling step-by-step! outperforming larger language models with less training data and smaller model sizes}.
\newblock In \emph{Findings of the Association for Computational Linguistics: ACL 2023}, pages 8003--8017, Toronto, Canada. Association for Computational Linguistics.

\bibitem[{Hu et~al.(2022)Hu, yelong shen, Wallis, Allen-Zhu, Li, Wang, Wang, and Chen}]{lora}
Edward~J Hu, yelong shen, Phillip Wallis, Zeyuan Allen-Zhu, Yuanzhi Li, Shean Wang, Lu~Wang, and Weizhu Chen. 2022.
\newblock \href {https://openreview.net/forum?id=nZeVKeeFYf9} {Lo{RA}: Low-rank adaptation of large language models}.
\newblock In \emph{International Conference on Learning Representations}.

\bibitem[{Hu et~al.(2017)Hu, Andreas, Rohrbach, Darrell, and Saenko}]{learning_to_reason}
Ronghang Hu, Jacob Andreas, Marcus Rohrbach, Trevor Darrell, and Kate Saenko. 2017.
\newblock Learning to reason: End-to-end module networks for visual question answering.
\newblock In \emph{Proceedings of the IEEE International Conference on Computer Vision (ICCV)}.

\bibitem[{Hudson and Manning(2019)}]{gqa}
Drew~A Hudson and Christopher~D Manning. 2019.
\newblock Gqa: A new dataset for real-world visual reasoning and compositional question answering.
\newblock In \emph{Proceedings of the IEEE/CVF conference on computer vision and pattern recognition}, pages 6700--6709.

\bibitem[{Hui et~al.(2024)Hui, Yang, Cui, Yang, Liu, Zhang, Liu, Zhang, Yu, Lu, Dang, Fan, Zhang, Yang, Men, Huang, Zheng, Miao, Quan, Feng, Ren, Ren, Zhou, and Lin}]{qwen25coder}
Binyuan Hui, Jian Yang, Zeyu Cui, Jiaxi Yang, Dayiheng Liu, Lei Zhang, Tianyu Liu, Jiajun Zhang, Bowen Yu, Keming Lu, Kai Dang, Yang Fan, Yichang Zhang, An~Yang, Rui Men, Fei Huang, Bo~Zheng, Yibo Miao, Shanghaoran Quan, Yunlong Feng, Xingzhang Ren, Xuancheng Ren, Jingren Zhou, and Junyang Lin. 2024.
\newblock \href {https://arxiv.org/abs/2409.12186} {Qwen2.5-coder technical report}.
\newblock \emph{Preprint}, arXiv:2409.12186.

\bibitem[{Jiang et~al.(2024)Jiang, Dong, Wang, Fang, Shang, Li, Jin, and Jiao}]{self-plan-code}
Xue Jiang, Yihong Dong, Lecheng Wang, Zheng Fang, Qiwei Shang, Ge~Li, Zhi Jin, and Wenpin Jiao. 2024.
\newblock Self-planning code generation with large language models.
\newblock \emph{ACM Trans. Softw. Eng. Methodol.}

\bibitem[{Johnson et~al.(2017)Johnson, Hariharan, Van Der~Maaten, Hoffman, Fei-Fei, Lawrence~Zitnick, and Girshick}]{johnson2017inferring}
Justin Johnson, Bharath Hariharan, Laurens Van Der~Maaten, Judy Hoffman, Li~Fei-Fei, C~Lawrence~Zitnick, and Ross Girshick. 2017.
\newblock Inferring and executing programs for visual reasoning.
\newblock In \emph{Proceedings of the IEEE international conference on computer vision}, pages 2989--2998.

\bibitem[{Khan et~al.(2024)Khan, BG, Schulter, Fu, and Chandraker}]{visrep}
Zaid Khan, Vijay~Kumar BG, Samuel Schulter, Yun Fu, and Manmohan Chandraker. 2024.
\newblock Self-training large language models for improved visual program synthesis with visual reinforcement.
\newblock In \emph{Proceedings of the IEEE/CVF Conference on Computer Vision and Pattern Recognition}, pages 14344--14353.

\bibitem[{Lewis et~al.(2020{\natexlab{a}})Lewis, Liu, Goyal, Ghazvininejad, Mohamed, Levy, Stoyanov, and Zettlemoyer}]{bart}
Mike Lewis, Yinhan Liu, Naman Goyal, Marjan Ghazvininejad, Abdelrahman Mohamed, Omer Levy, Veselin Stoyanov, and Luke Zettlemoyer. 2020{\natexlab{a}}.
\newblock \href {https://doi.org/10.18653/v1/2020.acl-main.703} {{BART}: Denoising sequence-to-sequence pre-training for natural language generation, translation, and comprehension}.
\newblock In \emph{Proceedings of the 58th Annual Meeting of the Association for Computational Linguistics}, pages 7871--7880, Online. Association for Computational Linguistics.

\bibitem[{Lewis et~al.(2020{\natexlab{b}})Lewis, Perez, Piktus, Petroni, Karpukhin, Goyal, K{\"u}ttler, Lewis, Yih, Rockt{\"a}schel et~al.}]{rag}
Patrick Lewis, Ethan Perez, Aleksandra Piktus, Fabio Petroni, Vladimir Karpukhin, Naman Goyal, Heinrich K{\"u}ttler, Mike Lewis, Wen-tau Yih, Tim Rockt{\"a}schel, et~al. 2020{\natexlab{b}}.
\newblock Retrieval-augmented generation for knowledge-intensive nlp tasks.
\newblock \emph{Advances in Neural Information Processing Systems}, 33:9459--9474.

\bibitem[{Liang et~al.(2023)Liang, Huang, Xia, Xu, Hausman, Ichter, Florence, and Zeng}]{robotics_vp}
Jacky Liang, Wenlong Huang, Fei Xia, Peng Xu, Karol Hausman, Brian Ichter, Pete Florence, and Andy Zeng. 2023.
\newblock \href {https://doi.org/10.1109/ICRA48891.2023.10160591} {Code as policies: Language model programs for embodied control}.
\newblock In \emph{2023 IEEE International Conference on Robotics and Automation (ICRA)}, pages 9493--9500.

\bibitem[{Lu et~al.(2024)Lu, Peng, Cheng, Galley, Chang, Wu, Zhu, and Gao}]{chameleon}
Pan Lu, Baolin Peng, Hao Cheng, Michel Galley, Kai-Wei Chang, Ying~Nian Wu, Song-Chun Zhu, and Jianfeng Gao. 2024.
\newblock Chameleon: Plug-and-play compositional reasoning with large language models.
\newblock \emph{Advances in Neural Information Processing Systems}, 36.

\bibitem[{Min et~al.(2024)Min, Buch, Nagrani, Cho, and Schmid}]{video_vp}
Juhong Min, Shyamal Buch, Arsha Nagrani, Minsu Cho, and Cordelia Schmid. 2024.
\newblock Morevqa: Exploring modular reasoning models for video question answering.
\newblock In \emph{Proceedings of the IEEE/CVF Conference on Computer Vision and Pattern Recognition (CVPR)}, pages 13235--13245.

\bibitem[{Minderer et~al.(2024)Minderer, Gritsenko, and Houlsby}]{owlvit2}
Matthias Minderer, Alexey Gritsenko, and Neil Houlsby. 2024.
\newblock Scaling open-vocabulary object detection.
\newblock \emph{Advances in Neural Information Processing Systems}, 36.

\bibitem[{{Open AI}(2024)}]{gpt40mini}
{Open AI}. 2024.
\newblock \href {https://openai.com/index/gpt-4o-mini-advancing-cost-efficient-intelligence/} {Gpt-4o mini: advancing cost-efficient intelligence}.

\bibitem[{Panagopoulou et~al.(2025)Panagopoulou, Zhou, Savarese, Xiong, Callison-Burch, Yatskar, and Niebles}]{viunit}
Artemis Panagopoulou, Honglu Zhou, Silvio Savarese, Caiming Xiong, Chris Callison-Burch, Mark Yatskar, and Juan~Carlos Niebles. 2025.
\newblock Viunit: Visual unit tests for more robust visual programming.
\newblock In \emph{Proceedings of the IEEE/CVF Conference on Computer Vision and Pattern Recognition (CVPR)}, pages 24646--24656.

\bibitem[{Radford et~al.(2021)Radford, Kim, Hallacy, Ramesh, Goh, Agarwal, Sastry, Askell, Mishkin, Clark et~al.}]{clip}
Alec Radford, Jong~Wook Kim, Chris Hallacy, Aditya Ramesh, Gabriel Goh, Sandhini Agarwal, Girish Sastry, Amanda Askell, Pamela Mishkin, Jack Clark, et~al. 2021.
\newblock Learning transferable visual models from natural language supervision.
\newblock In \emph{International conference on machine learning}, pages 8748--8763. PmLR.

\bibitem[{Raffel et~al.(2020)Raffel, Shazeer, Roberts, Lee, Narang, Matena, Zhou, Li, and Liu}]{t5}
Colin Raffel, Noam Shazeer, Adam Roberts, Katherine Lee, Sharan Narang, Michael Matena, Yanqi Zhou, Wei Li, and Peter~J Liu. 2020.
\newblock Exploring the limits of transfer learning with a unified text-to-text transformer.
\newblock \emph{Journal of machine learning research}, 21(140):1--67.

\bibitem[{Reimers and Gurevych(2019)}]{sentence-bert}
Nils Reimers and Iryna Gurevych. 2019.
\newblock \href {http://arxiv.org/abs/1908.10084} {Sentence-bert: Sentence embeddings using siamese bert-networks}.
\newblock In \emph{Proceedings of the 2019 Conference on Empirical Methods in Natural Language Processing}. Association for Computational Linguistics.

\bibitem[{Sanh et~al.(2019)Sanh, Debut, Chaumond, and Wolf}]{distillbert}
Victor Sanh, Lysandre Debut, Julien Chaumond, and Thomas Wolf. 2019.
\newblock Distilbert, a distilled version of bert: smaller, faster, cheaper and lighter.
\newblock In \emph{NeurIPS}.

\bibitem[{Schick et~al.(2024)Schick, Dwivedi-Yu, Dess, Raileanu, Lomeli, Hambro, Zettlemoyer, Cancedda, and Scialom}]{toolformer}
Timo Schick, Jane Dwivedi-Yu, Roberto Dess, Roberta Raileanu, Maria Lomeli, Eric Hambro, Luke Zettlemoyer, Nicola Cancedda, and Thomas Scialom. 2024.
\newblock Toolformer: Language models can teach themselves to use tools.
\newblock \emph{Advances in Neural Information Processing Systems}, 36.

\bibitem[{Stani{\'c} et~al.(2024)Stani{\'c}, Caelles, and Tschannen}]{codebison}
Aleksandar Stani{\'c}, Sergi Caelles, and Michael Tschannen. 2024.
\newblock Towards truly zero-shot compositional visual reasoning with llms as programmers.
\newblock \emph{arXiv preprint arXiv:2401.01974}.

\bibitem[{SU et~al.(2023)SU, Kasai, Wu, Shi, Wang, Xin, Zhang, Ostendorf, Zettlemoyer, Smith, and Yu}]{selective-in-context}
Hongjin SU, Jungo Kasai, Chen~Henry Wu, Weijia Shi, Tianlu Wang, Jiayi Xin, Rui Zhang, Mari Ostendorf, Luke Zettlemoyer, Noah~A. Smith, and Tao Yu. 2023.
\newblock Selective annotation makes language models better few-shot learners.
\newblock In \emph{The Eleventh International Conference on Learning Representations}.

\bibitem[{Subramanian et~al.(2023)Subramanian, Narasimhan, Khangaonkar, Yang, Nagrani, Schmid, Zeng, Darrell, and Klein}]{codevqa}
Sanjay Subramanian, Medhini Narasimhan, Kushal Khangaonkar, Kevin Yang, Arsha Nagrani, Cordelia Schmid, Andy Zeng, Trevor Darrell, and Dan Klein. 2023.
\newblock \href {https://doi.org/10.18653/v1/2023.acl-short.65} {Modular visual question answering via code generation}.
\newblock In \emph{Proceedings of the 61st Annual Meeting of the Association for Computational Linguistics (Volume 2: Short Papers)}, pages 747--761, Toronto, Canada. Association for Computational Linguistics.

\bibitem[{Sun et~al.(2024)Sun, Lyu, Li, Wan, Zhang, Li, and Jin}]{code-plan}
Zhihong Sun, Chen Lyu, Bolun Li, Yao Wan, Hongyu Zhang, Ge~Li, and Zhi Jin. 2024.
\newblock Enhancing code generation performance of smaller models by distilling the reasoning ability of {LLM}s.
\newblock In \emph{LREC-COLING}.

\bibitem[{Sur{\'\i}s et~al.(2023)Sur{\'\i}s, Menon, and Vondrick}]{vipergpt}
D{\'\i}dac Sur{\'\i}s, Sachit Menon, and Carl Vondrick. 2023.
\newblock Vipergpt: Visual inference via python execution for reasoning.
\newblock In \emph{Proceedings of the IEEE/CVF International Conference on Computer Vision}, pages 11888--11898.

\bibitem[{Tang et~al.(2019)Tang, Lu, Liu, Mou, Vechtomova, and Lin}]{distill-augmentation}
Raphael Tang, Yao Lu, Linqing Liu, Lili Mou, Olga Vechtomova, and Jimmy~J. Lin. 2019.
\newblock \href {https://api.semanticscholar.org/CorpusID:85543565} {Distilling task-specific knowledge from bert into simple neural networks}.
\newblock \emph{ArXiv}, abs/1903.12136.

\bibitem[{Tao et~al.(2024)Tao, T~V, Shlapentokh-Rothman, Gupta, Ji, and Hoiem}]{webwise}
Heyi Tao, Sethuraman T~V, Michal Shlapentokh-Rothman, Tanmay Gupta, Heng Ji, and Derek Hoiem. 2024.
\newblock \href {https://doi.org/10.18653/v1/2024.findings-naacl.234} {{W}eb{WISE}: Unlocking web interface control for {LLM}s via sequential exploration}.
\newblock In \emph{Findings of the Association for Computational Linguistics: NAACL 2024}, pages 3693--3711, Mexico City, Mexico. Association for Computational Linguistics.

\bibitem[{Tong et~al.(2024)Tong, II, Wu, Woo, IYER, Akula, Yang, Yang, Middepogu, Wang, Pan, Fergus, LeCun, and Xie}]{cvbench}
Shengbang Tong, Ellis L~Brown II, Penghao Wu, Sanghyun Woo, ADITHYA~JAIRAM IYER, Sai~Charitha Akula, Shusheng Yang, Jihan Yang, Manoj Middepogu, Ziteng Wang, Xichen Pan, Rob Fergus, Yann LeCun, and Saining Xie. 2024.
\newblock \href {https://openreview.net/forum?id=Vi8AepAXGy} {Cambrian-1: A fully open, vision-centric exploration of multimodal {LLM}s}.
\newblock In \emph{The Thirty-eighth Annual Conference on Neural Information Processing Systems}.

\bibitem[{Touvron et~al.(2023)Touvron, Lavril, Izacard, Martinet, Lachaux, Lacroix, Rozi{\`e}re, Goyal, Hambro, Azhar et~al.}]{llama}
Hugo Touvron, Thibaut Lavril, Gautier Izacard, Xavier Martinet, Marie-Anne Lachaux, Timoth{\'e}e Lacroix, Baptiste Rozi{\`e}re, Naman Goyal, Eric Hambro, Faisal Azhar, et~al. 2023.
\newblock Llama: Open and efficient foundation language models.
\newblock \emph{arXiv preprint arXiv:2302.13971}.

\bibitem[{Wang et~al.(2023)Wang, Le, Gotmare, Bui, Li, and Hoi}]{codet5plus}
Yue Wang, Hung Le, Akhilesh~Deepak Gotmare, Nghi~D.Q. Bui, Junnan Li, and Steven C.~H. Hoi. 2023.
\newblock Codet5+: Open code large language models for code understanding and generation.
\newblock \emph{arXiv preprint}.

\bibitem[{Wei et~al.(2022)Wei, Wang, Schuurmans, Bosma, ichter, Xia, Chi, Le, and Zhou}]{cot}
Jason Wei, Xuezhi Wang, Dale Schuurmans, Maarten Bosma, brian ichter, Fei Xia, Ed~Chi, Quoc~V Le, and Denny Zhou. 2022.
\newblock Chain-of-thought prompting elicits reasoning in large language models.
\newblock In \emph{Advances in Neural Information Processing Systems}.

\bibitem[{West et~al.(2021)West, Bhagavatula, Hessel, Hwang, Jiang, Bras, Lu, Welleck, and Choi}]{commonsense_distill}
Peter West, Chandrasekhar Bhagavatula, Jack Hessel, Jena~D. Hwang, Liwei Jiang, Ronan~Le Bras, Ximing Lu, Sean Welleck, and Yejin Choi. 2021.
\newblock \href {https://api.semanticscholar.org/CorpusID:238857304} {Symbolic knowledge distillation: from general language models to commonsense models}.
\newblock In \emph{North American Chapter of the Association for Computational Linguistics}.

\bibitem[{Wu et~al.(2024)Wu, Lin, Zhao, Wu, Lu, Peng, and Chang}]{vdebugger}
Xueqing Wu, Zongyu Lin, Songyan Zhao, Te-Lin Wu, Pan Lu, Nanyun Peng, and Kai-Wei Chang. 2024.
\newblock \href {https://doi.org/10.18653/v1/2024.findings-emnlp.575} {{VD}ebugger: Harnessing execution feedback for debugging visual programs}.
\newblock In \emph{Findings of the Association for Computational Linguistics: EMNLP 2024}, pages 9845--9860, Miami, Florida, USA. Association for Computational Linguistics.

\bibitem[{Yuan et~al.(2024)Yuan, Chen, Wang, Fung, Peng, and Ji}]{craft}
Lifan Yuan, Yangyi Chen, Xingyao Wang, Yi~Fung, Hao Peng, and Heng Ji. 2024.
\newblock \href {https://openreview.net/forum?id=G0vdDSt9XM} {{CRAFT}: Customizing {LLM}s by creating and retrieving from specialized toolsets}.
\newblock In \emph{The Twelfth International Conference on Learning Representations}.

\bibitem[{Zhang et~al.(2018)Zhang, Galley, Gao, Gan, Li, Brockett, and Dolan}]{ngramentropy}
Yizhe Zhang, Michel Galley, Jianfeng Gao, Zhe Gan, Xiujun Li, Chris Brockett, and Bill Dolan. 2018.
\newblock Generating informative and diverse conversational responses via adversarial information maximization.
\newblock \emph{Advances in Neural Information Processing Systems}, 31.

\bibitem[{Zhang et~al.(2023)Zhang, Zhang, Li, and Smola}]{auto-cot}
Zhuosheng Zhang, Aston Zhang, Mu~Li, and Alex Smola. 2023.
\newblock Automatic chain of thought prompting in large language models.
\newblock In \emph{The Eleventh International Conference on Learning Representations}.

\bibitem[{Zhu et~al.(2023)Zhu, Qi, Zhang, Long, and Zhou}]{program_distillation}
Xuekai Zhu, Biqing Qi, Kaiyan Zhang, Xingwei Long, and Bowen Zhou. 2023.
\newblock \href {https://api.semanticscholar.org/CorpusID:258840866} {Pad: Program-aided distillation can teach small models reasoning better than chain-of-thought fine-tuning}.
\newblock In \emph{North American Chapter of the Association for Computational Linguistics}.

\end{thebibliography}

\appendix

\appendix 
\section{Additional Results}
\label{sec:add_results}
Additional experimental results on two vision-centric datasets, the 2D-split of CV Bench~\cite{cvbench} and the VizWiz~\cite{vizwiz} validation set, can be found in Table~\ref{tab:additional_dataset}. We evaluate an augmented student model (Qwen2.5-Coder-0.5B) and the teacher model. For CV Bench~\cite{cvbench}, the student model is trained on both GQA and VQAv2 (referred to as `joint' in Section~\ref{sec:exp}). 2000 examples (~0.1\%) of the VizWiz training set are annotated by the teacher model and used to train the student. Like the experiments on VQAv2, we use BERT~\cite{bert} and BART~\cite{bart}, for augmentation. The results show the efficacy of our method on datasets related to the training set as well as real-world tasks. 
\begin{table}[!htb]
  \centering
  \scriptsize
      \resizebox{\columnwidth}{!}{
  \setlength{\tabcolsep}{4pt}
  \begin{tabular}{lcccc}
    \toprule
    Method 
      & \multicolumn{2}{c}{\textbf{CV Bench}} 
      & \multicolumn{2}{c}{\textbf{VizWiz}} \\
    \cmidrule(lr){2-3} \cmidrule(lr){4-5}
      & Answer Acc & Stu/Tea Agree. 
      & Answer Acc & Stu/Tea Agree. \\
    \midrule
    GPT-4o-Mini (Teacher)
      & 56.0   & N/A          
      &    34.3   &   N/A    \\
    Qwen Coder\,2.5-0.5B (w/ Aug)  
      &   54.5    & 93.0           
      &     34.9  &     71.6      \\
    \bottomrule
  \end{tabular}}
  \caption{Additional evaluation of augmented distillation on the 2D split of CV Bench and the validation split of the real-world dataset VizWiz. The distilled augmented model has similar performance as the teacher model.}
  \label{tab:additional_dataset}
\end{table}

\paragraph{N-Gram Entropy} In our experiments, we primarily use task-level metrics to validate our method and assess augmented data quality. To directly quantify data diversity, we also measure N-Gram Entropy \cite{ngramentropy} on 20,000 samples from both our augmented and non-augmented GQA training sets. The augmented data has an entropy of 7.10 compared to 6.18 for the original set, confirming that our augmentation procedure meaningfully increases linguistic variety.

\paragraph{Probability of Augmentation} In Section~\ref{sec:aug}, we state that during augmentation, each argument has a 50\% probability of being replaced. We evaluate using different percentages on a small validation set of GQA and found 50\% had the highest performance, 43.2\%,  compared to 42.5\% for 90\% replacement rate and 42.4\% for a 10\% replacement rate. 

\paragraph{Model Size} The focus of our work is to design a method for a small amount of training data as well small models that can be run on consumer GPUs, which led us to focus on models with no more than 1B parameters. However, there are models that are slightly larger, such as 1.5B  which leads to the question,  whether there are any performance gains from using such models. On a subset of GQA, we evaluate different sizes of Qwen2.5-Coder on 0.5B, 1.5B and 3B. The results, shown in Table~\ref{tab:model_size}, show our augmentation method is actually more effective on smaller models. Scaling our method for larger open source models is an area for future work. 
\begin{table}[!htb]
  \centering
  \scriptsize
  \setlength{\tabcolsep}{4pt}
  \begin{tabular}{lcc}
    \toprule
    Method 
      & \multicolumn{2}{c}{\textbf{GQA Subset}} 
    \\
    \cmidrule(lr){2-3}
      & Answer Acc & Stu/Tea Agree.  \\
    \midrule
    Qwen Coder\,2.5-0.5B (w/Aug)
      & \textbf{43.2}   & \textbf{70.5}                 \\
    Qwen Coder\,2.5-1.5B(w/ Aug)  
      & 42.5       & 65.8     \\
     Qwen Coder\,2.5-3B(w/ Aug) 
     & 42.4 & 67.6 \\
    \bottomrule
  \end{tabular}
  \caption{When using $0.1\%$ of the training dataset, our method is more effective on smaller models,}
  \label{tab:model_size}
\end{table}

\section{Program Analysis}
\label{sec:program-analysis}
\begin{figure*}
    \centering
    \includegraphics[width=1\linewidth]{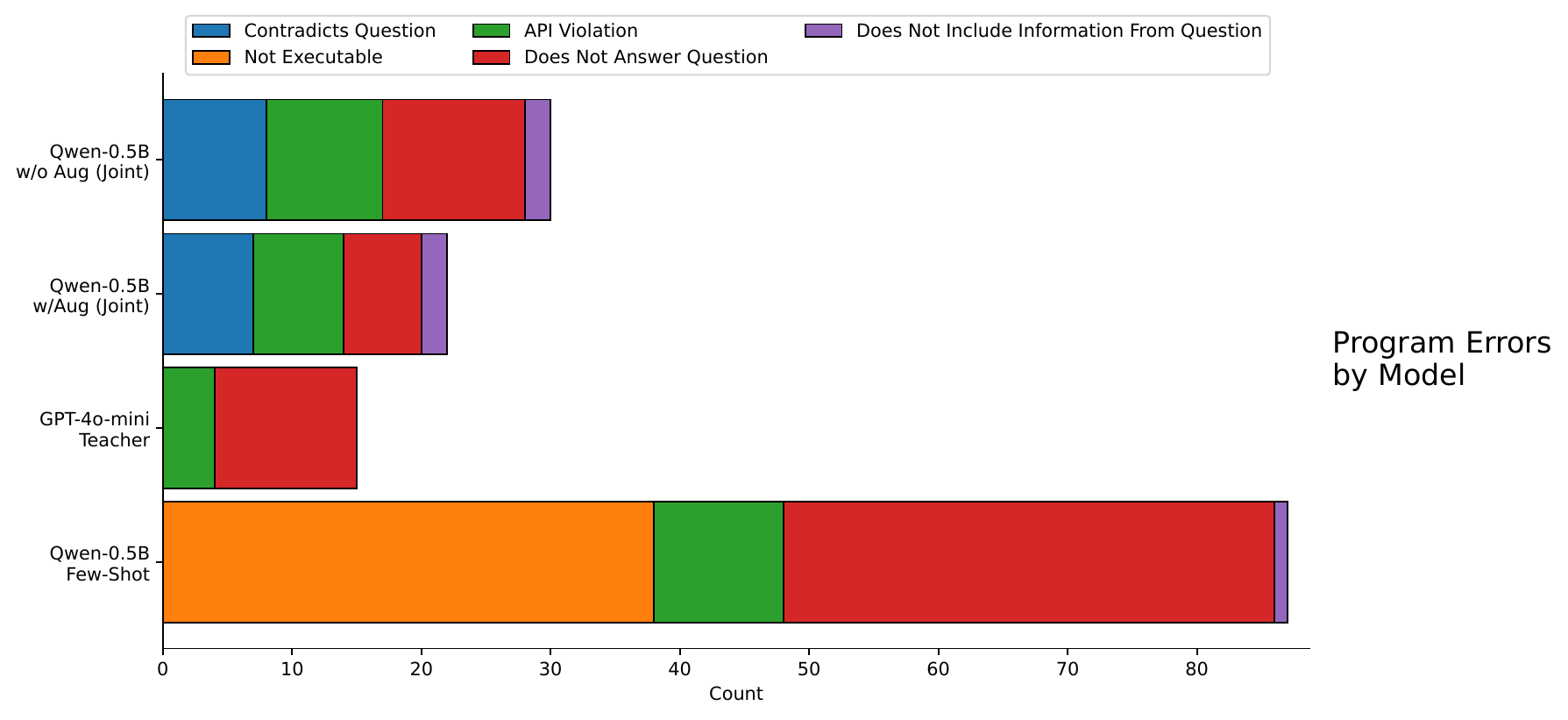}
    \caption{The frequency of errors across the different categories for GQA program evaluation. Augmentation reduces the number of `Does Not Answer Question Mistakes.' }
    \label{fig:bar-char}
\end{figure*}

\begin{figure*}
    \centering
    \includegraphics[width=1\linewidth]{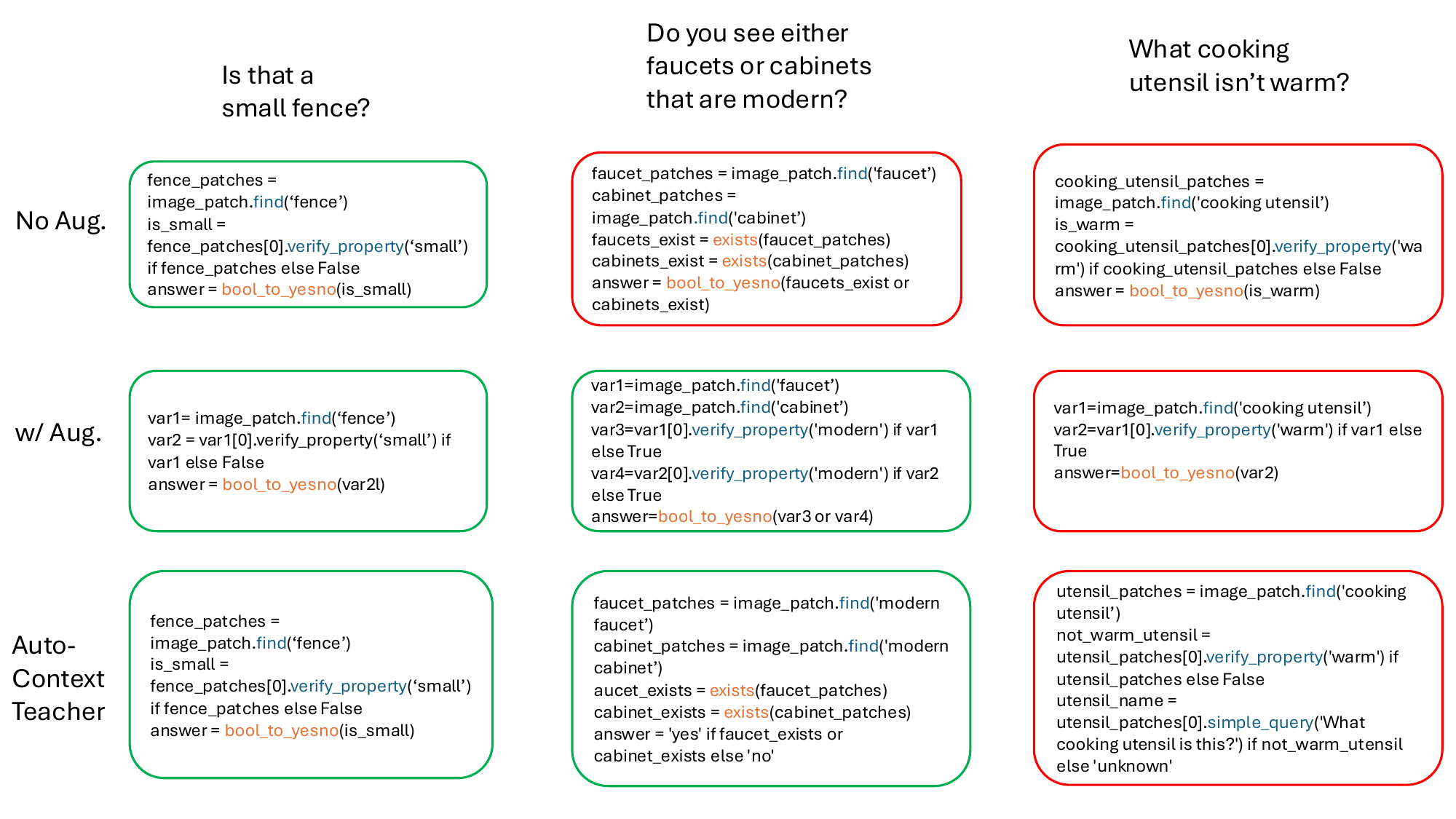}
    \caption{3 question/programs using no augmentation, augmentation and auto-context teacher. Simple comparison questions (left hand side) are almost always correct while questions with negations are almost always incorrect across the different methods.  }
    \label{fig:qualitative}
\end{figure*}

\paragraph{Program Errors} In Figure~\ref{fig:bar-char}, we show the frequency of different error types for Qwen-Coder-0.5B under few-shot model, non-augmented (joint) model and augmented (joint) model  as well as the teacher model on the GQA dataset. Note that incorrect programs can fall into multiple categories but in this classification, each incorrect program was counted only once. Other than the few-shot model, there are no execution errors. The main sources of error are not answering question and API violation. 
\paragraph{Qualitative Analysis}  In Figure~\ref{fig:qualitative}, we show 9 generated programs from  3 questions in the GQA dataset and 3 models: CodeT5 student model trained without augmentation, CodeT5 with augmentation and the auto-context teacher. Generated programs for existence questions about a single object and a single attribute like the one in the first column are almost always correct, even for few-shot methods. In the second column, we see an example where the model trained without augmentation leaves out details mentioned in the question  but the augmented model generates the correct program. All of the programs are incorrect in the last column but for different reasons. Both of the student models use the same program (apart from variable names) and make two mistakes.  First the question asks about `not warm' instead of `warm' and the second is that the answer should be an object, not yes/no. The teacher program returns an object but still fails to recognize that the object should be `not warm' even though the variable name includes `not warm' in it. Questions involving negative properties are almost always missed by the teacher and student models. 

\section{Training and Model Details}
\label{sec:training}
We used the following models for executing programs:
\begin{enumerate}
    \item CLIP ViT-L/14~\cite{clip}
    \item InstructBLIP Flan-T5 XL~\cite{instructblip}
    \item OWLv2 Base Patch 16 Ensemble~\cite{owlvit2}
\end{enumerate}
Program generation settings for GPT can be found in Table~\ref{tab:gpt_hyper}. Training hyper-parameters can be found in Table~\ref{tab:template_based_train}. For distilled models, the most important hyper-parameters were the learning rate and LoRA dropout rate. Training stopped when the training loss stopped decreasing. 

All experiments were run on a single 40gb A40 or 40gb A100. Time measurements were taken on an A40. 
\begin{table}[]
    \centering
    \begin{tabular}{cc}
        \toprule
         Setting & Value \\ 
         \midrule 
         Temperature& 0   \\
         
         Top\_p & 1.0 \\ 
         Frequency Penalty & 0.0 \\ 
         Presence Penalty & 0.0 \\
         Max Output Tokens & 256\\
         \bottomrule
    \end{tabular}
    \caption{GPT-4o-mini generation settings}
    \label{tab:gpt_hyper}
\end{table}
\begin{table}[]
\resizebox{\columnwidth}{!}{
    \centering
    \begin{tabular}{cc}
    \toprule
         Hyper-parameters & Value \\
         \midrule 
         LoRA target modules & All linear layers  \\
        LoRA rank &  8 \\ 
        LoRA alpha & 16\\
        LoRA bias & None \\
        LoRA dropout & 0.05 (no augmentation), 0.1 (augmentation)\\
        LR & 2e-4 \\
        Batch Size  & 16 \\ 
        Max Output Tokens & 256\\
        \bottomrule
    \end{tabular}}
    \caption{Training and evaluation settings for student models. We use the same learning rate for all models.}
    \label{tab:template_based_train}
\end{table}
\section{Changes to ViperGPT API}
\label{sec:viperapi}
The following are major modifications made to the ViperGPT API~\cite{vipergpt}. 
\begin{figure}
    \centering
    \includegraphics[width=1\linewidth]{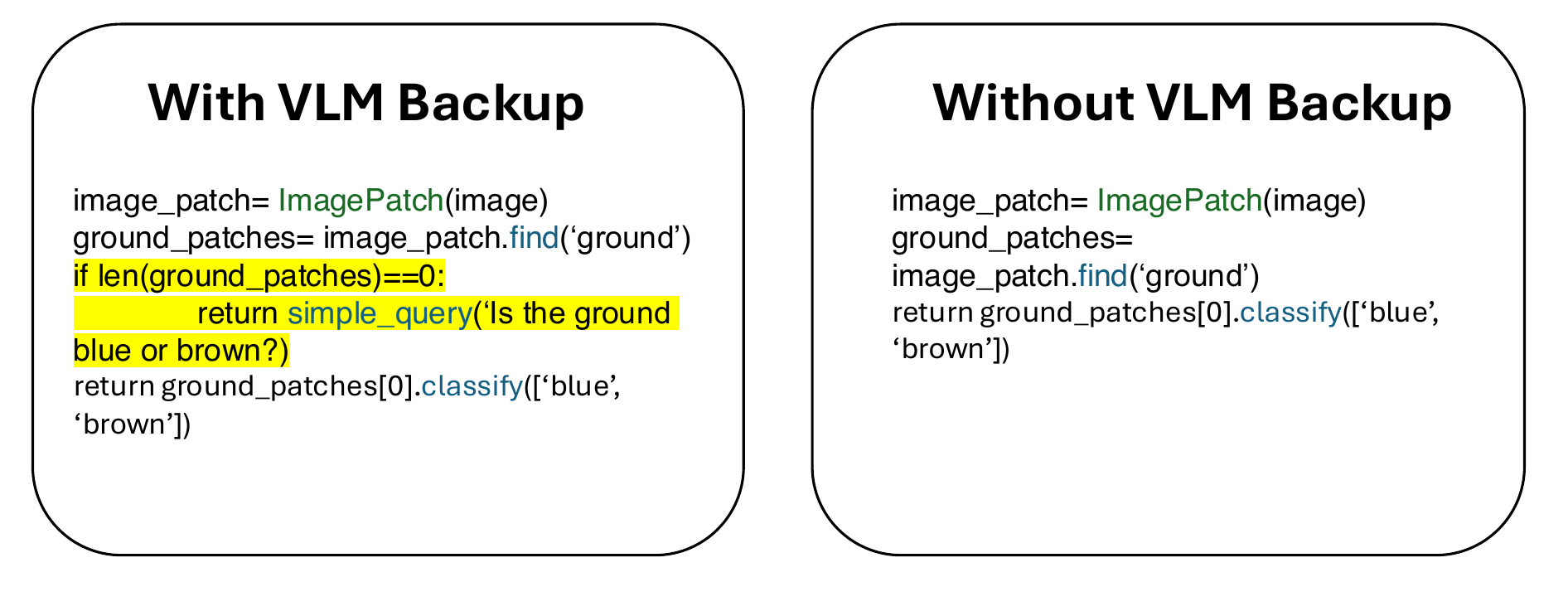}
    \caption{Difference in program annotations when a VLM is used as a backup model for the question `Is the ground blue or brown?' The highlighted portion is removed from all program annotations used.  }
    \label{fig:vlm_backup}
\end{figure}
\begin{enumerate}
    \item Program annotations were modified not to use a vision-language model (VLM) when the program fails (see Figure~\ref{fig:vlm_backup} for an example). In the original ViperGPT API, examples in the API included a line to directly query a VLM if other parts of the program failed such as when no object is found. The performance using the original ViperGPT code decreases considerably when the VLM backup lines are removed from the API as shown in Table~\ref{tab:vipergpt_vlm}. 
    \item An object is always returned by the object detector. 
    \item Program annotations did not include parts of the program that were shared among all examples. 
    \item Several new functions were added.  
    \begin{enumerate}
        \item Verify Relationship: Given two objects and a relation, return a boolean whether the objects satisfy that relationship.
        \item Choose Relationship: Given two objects, return the relationship between the two. 
        \item Crop Position: Crop part of the image based on a position relative to an object
    \end{enumerate}
\end{enumerate}

\begin{table}[]
    \centering
    \resizebox{\linewidth}{!}{ 
    \begin{tabular}{cc}
    \toprule
        Use of VLM Backup         & GQA-Test Dev \\ 
        \midrule 
       ViperGPT with VLM Backup  &  47.3\\
         ViperGPT without VLM Backup &  26.0\\
         \bottomrule
    \end{tabular}}
    \caption{Change in GQA test-dev accuracy using original ViperGPT API when not using a VLM when the program fails }
    \label{tab:vipergpt_vlm}
\end{table}

\onecolumn

\onecolumn

\label{sec:prompt_ours}

\lstset{
    basicstyle=\ttfamily\footnotesize,
    breaklines=true,
    breakatwhitespace=true,
    frame=none,
    numbers=none,
    stepnumber=1,
    showstringspaces=false
}

\onecolumn
\section{Prompt}
\label{sec:prompt_ours}

\begin{lstlisting}[language=Python]
Instructions
---------------
For each question provided, generate a Python program that includes a return statement. Assume that image_patch = ImagePatch(image) is already defined. The final output of the program should always be a string.
---------------
ImagePatch
---------------
Attributes
---------------
1. **cropped_image**
    Type: array
    Description: An array representing the cropped image.
2. **left**
    Type: int
   Description: The left border of the crop's bounding box.
3. **lower**
    Type: int
    Description: The bottom border of the crop's bounding box.
4. **right**
    Type: int
    Description: The right border of the crop's bounding box.
5. **upper**
    Type: int
    Description: The top border of the crop's bounding box.
Methods
---------------
1. **find(object_name: str) -> List[ImagePatch]**
    Description: Returns a list of image patches containing the specified object.
    Notes: find should not be the last operation in a program.
    Examples:
    image_patch.find('chair')
    image_patch.find('table')

2. **crop_position(direction: str, reference_patch: ImagePatch) -> ImagePatch**
    Description: Returns a new image patch in the specified direction relative to the reference_patch. Directions can include 'left', 'right', 'above', 'below', 'on', 'in front', etc.
    Notes: The result of crop_position should not be immediately indexed on the next line. The second argument is always the original image_patch. 
    Examples:
    image_patch.crop_position('left', image_patch)
    image_patch.crop_position('above', image_patch)

3. **verify_property(property_name: str) -> bool**
    Description: Returns True if the object contains the specified property; otherwise, False.
    Notes: Can only be called on an image patch.
    Examples:
    image_patch.verify_property('red')
    image_patch.verify_property('running')

4. **classify(options: Union[str, List[str]]) -> str**
    Description: Given a category (e.g., 'color', 'material', 'furniture') or a list of options, returns the best option for the image patch.
    Notes: The input should not be 'object'.
    Examples:
    image_patch.classify(['red', 'blue'])
    image_patch.classify('color')

5. **simple_query(question: str) -> str**
    Description: Answers questions about the image, especially ambiguous ones (e.g., 'Who is riding?').
    Examples:
    image_patch.simple_query('Who is riding?')

General Functions
---------------
1. **filter_img(image_patches: List[ImagePatch], criteria: str) -> List[ImagePatch]**
    Description: Filters the list of image patches based on the given criteria. The criteria can be an action, attribute, or object.
    Examples:
    filter_img(image_patches, 'red')
    filter_img(image_patches, 'running')

2. **choose_relationship(patch1: Union[ImagePatch, List[ImagePatch]], patch2: Union[ImagePatch, List[ImagePatch]], relationships: Union[List[str], str]) -> str**
    Description: Chooses the relationship that best matches the two patches from the provided options.
    Examples:
    choose_relationship(image_patch1, image_patch2, ['on top of', 'next to'])
    choose_relationship(image_patch1, image_patch2, ['left', 'right'])

3. **verify_relationship(patch1: Union[ImagePatch, List[ImagePatch]], patch2: Union[ImagePatch, List[ImagePatch]], relationship: str) -> str**
    Description: Returns 'yes' or 'no' based on whether the specified relationship holds between the two patches.
    Examples:
    verify_relationship(image_patch1, image_patch2, 'on top of')
    verify_relationship(image_patch1, image_patch2, 'left')
4. **exists(patches: Union[ImagePatch, List[ImagePatch]]) -> bool**
    Description: Checks whether any of the provided image patches exist.
    Notes: If used as the last operation, it should be followed by bool_to_yesno().
    Examples:
    exists(image_patches)

5. **bool_to_yesno(value: bool) -> str**
    Description: Converts a boolean value to 'yes' or 'no'. Used to convert outputs of verify_property and exists.
    Examples:
    bool_to_yesno(exists(image_patches))
Here are some examples of how to write programs:

{examples}
Additional Notes
---------------
- You may utilize standard Python functions within your programs.
- Do not include comments.
- Only return the program.
- Do not define the function.
- Functions never return None.
- The last line of each program should be answer = 
\end{lstlisting}

\section{Variable Renamer}
\label{sec:rename}
\label{sec:var_renamer}
\lstset{
    basicstyle=\ttfamily\footnotesize,
    breaklines=true,
    breakatwhitespace=true,
    frame=none,
    numbers=none,
    stepnumber=1,
    showstringspaces=false
}

\begin{lstlisting}[language=Python]
class VariableRenamer(ast.NodeTransformer):
    def __init__(self, skip_vars=None):
        self.counter = 1                # For general variables (var1, var2, ...)
        self.temp_counter = 1           # For comprehension/loop variables (temp_var_1, ...)
        self.name_map = {}
        self.skip_vars = set(skip_vars) if skip_vars else set()

    def _new_name(self):
        name = f"var{self.counter}"
        self.counter += 1
        return name

    def _new_temp_name(self):
        name = f"temp_var_{self.temp_counter}"
        self.temp_counter += 1
        return name

    def rename_target(self, target):
        """Rename normal assignment or loop targets, skipping those in skip_vars."""
        if isinstance(target, ast.Name):
            if target.id in self.skip_vars:
                return target
            if target.id not in self.name_map:
                self.name_map[target.id] = self._new_name()
            target.id = self.name_map[target.id]
        elif isinstance(target, (ast.Tuple, ast.List)):
            for elt in target.elts:
                self.rename_target(elt)
        return target

    def visit_Name(self, node):
        if isinstance(node.ctx, (ast.Store, ast.Load, ast.Del)):
            if node.id in self.skip_vars:
                return node
            if node.id in self.name_map:
                node.id = self.name_map[node.id]
        return node

    def visit_Assign(self, node):
        node.value = self.visit(node.value)
        node.targets = [self.rename_target(t) for t in node.targets]
        return node

    def rename_within(self, node, old_name, new_name):
        """Recursively replace occurrences of old_name with new_name within the node."""
        class NameReplacer(ast.NodeTransformer):
            def visit_Name(self, n):
                if n.id == old_name:
                    n.id = new_name
                return n
        replacer = NameReplacer()
        return replacer.visit(node)

    def visit_For(self, node):
        # Enhanced handling for For loops to propagate renaming within the loop body.
        if isinstance(node.target, ast.Name) and node.target.id not in self.skip_vars:
            old_name = node.target.id
            new_temp = self._new_temp_name()
            node.target.id = new_temp

            # Visit and rename within 'iter', 'body', and 'orelse'
            node.iter = self.visit(node.iter)
            node.body = [self.rename_within(self.visit(n), old_name, new_temp) for n in node.body]
            if node.orelse:
                node.orelse = [self.rename_within(self.visit(n), old_name, new_temp) for n in node.orelse]
        else:
            node.target = self.rename_target(node.target)
            node.iter = self.visit(node.iter)
            node.body = [self.visit(n) for n in node.body]
            if node.orelse:
                node.orelse = [self.visit(n) for n in node.orelse]
        return node

    def visit_While(self, node):
        node.test = self.visit(node.test)
        node.body = [self.visit(n) for n in node.body]
        if node.orelse:
            node.orelse = [self.visit(n) for n in node.orelse]
        return node

    def visit_ListComp(self, node):
        for gen in node.generators:
            if isinstance(gen.target, ast.Name) and gen.target.id not in self.skip_vars:
                old_name = gen.target.id
                new_temp = self._new_temp_name()
                gen.target.id = new_temp

                node.elt = self.rename_within(node.elt, old_name, new_temp)
                gen.ifs = [self.rename_within(if_clause, old_name, new_temp) for if_clause in gen.ifs]
                for inner_gen in node.generators:
                    inner_gen.target = self.rename_within(inner_gen.target, old_name, new_temp)
            else:
                gen.target = self.rename_target(gen.target)
            gen.iter = self.visit(gen.iter)
        node.elt = self.visit(node.elt)
        for gen in node.generators:
            gen.ifs = [self.visit(if_clause) for if_clause in gen.ifs]
        return node

    def visit_GeneratorExp(self, node):
        for gen in node.generators:
            if isinstance(gen.target, ast.Name) and gen.target.id not in self.skip_vars:
                old_name = gen.target.id
                new_temp = self._new_temp_name()
                gen.target.id = new_temp

                node.elt = self.rename_within(node.elt, old_name, new_temp)
                gen.ifs = [self.rename_within(if_clause, old_name, new_temp) for if_clause in gen.ifs]
                for inner_gen in node.generators:
                    inner_gen.target = self.rename_within(inner_gen.target, old_name, new_temp)
            else:
                gen.target = self.rename_target(gen.target)
            gen.iter = self.visit(gen.iter)
        node.elt = self.visit(node.elt)
        for gen in node.generators:
            gen.ifs = [self.visit(if_clause) for if_clause in gen.ifs]
        return node

    def visit_With(self, node):
        for item in node.items:
            if item.optional_vars and isinstance(item.optional_vars, ast.Name) and item.optional_vars.id not in self.skip_vars:
                item.optional_vars.id = self._new_temp_name()
            elif item.optional_vars:
                item.optional_vars = self.rename_target(item.optional_vars)
            item.context_expr = self.visit(item.context_expr)
        node.body = [self.visit(n) for n in node.body]
        return node

    # Additional visitor methods for other constructs can be added here.
def format_assignments(source_code: str) -> str:
    """
    Remove spaces around the equals sign in single-line assignment statements
    without altering multi-line assignments.
    
    This function ensures that:
    - Single-line assignments have no spaces around '='.
    - Multi-line assignments are left intact to preserve code correctness.
    """
    lines = source_code.split('\n')
    formatted_lines = []
    assignment_pattern = re.compile(r'^(\s*)(\w+)\s*=\s*(.+)$')

    # Track the balance of parentheses, brackets, and braces
    paren_balance = 0

    for line in lines:
        stripped_line = line.strip()

        # Update paren_balance
        paren_balance += line.count('(') - line.count(')')
        paren_balance += line.count('[') - line.count(']')
        paren_balance += line.count('{') - line.count('}')

        # If paren_balance > 0, we're inside a multi-line expression
        if paren_balance > 0:
            formatted_lines.append(line)
            continue

        # Attempt to match an assignment statement
        match = assignment_pattern.match(line)
        if match:
            indent, var, expr = match.groups()
            # Remove spaces around '=' and reconstruct the line
            formatted_line = f"{indent}{var}={expr}"
            formatted_lines.append(formatted_line)
        else:
            # Non-assignment lines are added directly
            formatted_lines.append(line)

    # Join the lines back into a single string
    return '\n'.join(formatted_lines)


def replace_variables(code: str, convert_to_source: bool = True) -> Union[str, ast.AST]:
    skip_list = {"image_patch", "answer"}  # Variables not to rename

    tree = ast.parse(code)
    renamer = VariableRenamer(skip_vars=skip_list)
    new_tree = renamer.visit(tree)
    ast.fix_missing_locations(new_tree)
    new_source = ast.unparse(new_tree)
    formatted_source = format_assignments(new_source)
    return formatted_source


\end{lstlisting}

Given a generated program we call the function `replace\_variables' which uses and abstract-syntax tree to rename variables both in and outside different types of loops. 
\twocolumn 
\section{Variable Replacement}
\label{sec:replace}
\begin{algorithm}
\caption{Argument Replacement}
\label{alg:replace}
\begin{algorithmic}
\STATE Extract arguments per function
\FOR {func in functions}
    \FOR {arg in arguments}
        \IF {arg in category}
            \STATE {Random sample from category}
        \ELSE
            \STATE {Random sample from generic object list} 
        \ENDIF 
    \ENDFOR
\ENDFOR

\end{algorithmic}
\end{algorithm}
The general algorithm for replacing an argument in a program can be seen in Algorithm~\ref{alg:replace}. For each named function or method in the API, we extract the arguments. For GQA, if the argument is already in a pre-defined catgory, we randomly sample from that category. Otherwise we randomly sample an object. Some example categories and options can be seen in Table~\ref{tab:categories}. The process for VQA is similar except there is no pre-defined list. Instead we mask out the argument in the question and generate a replacement using BERT or BART if the argument is a phrase. For full questions, we use word tokenization (default NLTK Tokenization) and POS-tagging with NLTK~\cite{nltk} to determine where to place masks. We randomly sample from the top-50 results. Both the BERT and BART models are large uncased with 340 and 406 M parameters.  

We perform the process above if an argument is selected for replacement during training. Each argument has a probability $p=0.5$ to be selected.

\begin{table}[h]
\centering

\caption{Argument Categories and Options}
\label{tab:categories}
\resizebox{\linewidth}{!}{ 
\begin{tabular}{ll} 
\toprule
\textbf{Category Name} & \textbf{Category Examples} \\
\midrule
Color & \makecell{red, blue, green, yellow,\\ purple, black, white,\\ orange, pink, brown, \\gray, indigo, cyan \\ magenta, tan, silver} \\
\midrule
Activities & \makecell{running, walking, snowboarding, \\ flying, splashing, tossing, \\ riding, standing,
hugging \\ hanging, breaking, pulling, \\ decorating, facing, preparing \\ pouring, pointing, laughing }\\ 
\midrule 
Relation & \makecell{picking up, in front of\\ behind, above, below \\ next to, near, far away \\ close , following, on top, \\ beside, walking on, attached, \\ left, right, diagonal} \\
\bottomrule
\end{tabular}}
\end{table}

\end{document}